\documentclass{article}

\usepackage[preprint]{tmlr}

\usepackage[utf8]{inputenc}
\usepackage[T1]{fontenc}
\usepackage{hyperref}
\usepackage{url}
\usepackage{booktabs}
\usepackage{longtable}
\usepackage{amsfonts}
\usepackage{amsmath}
\usepackage{microtype}
\usepackage{xcolor}
\usepackage{graphicx}
\usepackage{array}
\usepackage[font=small,labelfont=bf]{caption}

\graphicspath{{./}}

\hypersetup{colorlinks=true, linkcolor=black, citecolor=black, urlcolor=blue!50!black}

\newcommand{\toolname}{\textsc{OpenGameEval}}
\newcommand{\code}[1]{\texttt{\small #1}}

\newcommand{\repo}{\url{https://github.com/Roblox/open-game-eval}}
\newcommand{\mcpdocs}{\url{https://create.roblox.com/docs/studio/mcp}}
\newcommand{\editor}{Roblox Studio}
\newcommand{\plat}{Roblox}
\newcommand{\lang}{Luau}

\title{\toolname{}: Benchmarking Agentic Programming\\
and Exploration in a Stateful Game Engine%
\thanks{Large language model tools assisted with software development, data-analysis
implementation, and manuscript editing. The authors directed the research and verified all
code, results, and claims.}}

\author{%
  Eray Turkel\thanks{Corresponding author: \texttt{eturkel@alumni.stanford.edu}.}
  \quad Mengsha Sun \quad Kartik Ayyar \quad Sean Dunigan \quad Jack Lu\\
  Vlad Shcherban \quad Hsiang-Shun Shih \quad Xin Wang \quad Tiantian Zhang\\
  {\addr Roblox}
}

\begin{document}

\maketitle

\begin{abstract}
We present \toolname{}, a benchmark and evaluation framework for agentic game
development inside \editor{}. It runs language models as agents in reproducible, stateful
game-engine sessions and scores each run with executable checks, both on the edited scene and in
a simulated play session. Most agentic coding benchmarks require exploration but score only
final task success. \toolname{} separates observation tools from editing tools in its eight-tool
action space, so exploration can be measured directly. We measure the pass rates and exploration
behavior of $13$ frontier models on $84$ human-curated core tasks, with $16$ attempts per task. 

The tasks are hard for current models. The best model solves $51.7\%$ of tasks on a single
attempt and $39.4\%$ five times out of five, and no tested model solves six of the tasks. Models
at the frontier reach similar pass rates by solving different tasks: splitting tasks by the kind
of work they require spreads the top five by $5.0$pp on script-authoring tasks and $12.5$pp on
scene-change tasks.

Exploration behavior predicts whether a run succeeds. Holding task and model fixed, a run
that inspects every object a reference solution touches \emph{before acting on it} passes
$13.4$pp more often than a run that inspects none of them on scene-only tasks, and $9.8$pp more
often on script-only tasks.

We release the task suite, its place files, the per-task annotations, a plugin that runs the
tasks inside \editor{}, and an updated leaderboard under the MIT license at \repo{}.
\end{abstract}

\section{Introduction}
\label{sec:intro}

Code-generating language models are mostly evaluated on outcomes: a model receives a
problem, and a fixed oracle checks its output. \textsc{HumanEval} and \textsc{MBPP} score
function bodies against unit tests \citep{chen2021codex,austin2021mbpp}, and \textsc{SWE-bench}
and its verified subset score patches against a repository's own test suite
\citep{jimenez2024swebench,chowdhury2024verified}. This approach has driven substantial
progress, but it assumes that the final outcome is the only quantity worth measuring. The same
passing result can come from an agent that understood the system or from one that guessed, and
the outcome alone cannot tell them apart \citep{zhu2025abc,bean2025constructvalidity}. We study
a domain where that assumption breaks down, and we focus on exploration: what a model finds out
about the world it is working in.

Consider an AI coding assistant inside a game engine that is asked to ``implement health
regeneration that begins two seconds after damage and restores ten health per second'' in an
existing game. The prompt does not say which scripts manage health, whether damage is
server-authoritative, or where in the object hierarchy the player rig lives. The agent has to
find these facts itself, by searching the object hierarchy, reading the scripts already in the
game, and inspecting the properties of the objects it finds. Correctness is also a play-time
property. It can only be observed once the game runs, the damage event fires, and the client
sees health climb on the right schedule.

We introduce \toolname{}, a benchmark and evaluation framework for agentic game development
that runs an LLM agent inside a reproducible editor session. Four design properties together
let us measure exploration, not just outcomes: (i) a \textbf{stateful world with hidden
structure} that the agent must discover through tools (Figure~\ref{fig:env}), (ii) an
\textbf{instrumented eight-tool action space} that separates observation from action, (iii)
\textbf{play-time verification} with real physics, networking, and client--server dynamics, and
(iv) \textbf{human-written reference solutions} annotated with the objects and scripts a correct
run should consult. We evaluate 13 models on the 84-task core suite, with 16 attempts per task,
and report three main results.

First, the tasks are hard for current models. The strongest model solves $51.7\%$ of core
tasks on a single attempt and only $39.4\%$ five times out of five. No model solves six of the
core tasks in any of their 16 attempts, although each task's human-written reference solution
passes its checks.

Second, the models at the frontier do not fall into a simple order. Five models finish within
$1.1$pp of one another at about $51\%$, but they get there by solving different tasks. All five
solve only 27 of the 84 core tasks, while at least one of them solves $58$
(\S\ref{sec:frontier}). Scoring each kind of work separately shows where they differ: the spread
among the same five models grows from $1.1$pp to $5.0$pp on script-authoring tasks and to
$12.5$pp on scene-change tasks (Figure~\ref{fig:kinds}). A single average pass rate hides how
different these models are.

Third, how a run explores predicts whether it succeeds. The first two results concern which
tasks a model solves, and this one concerns the trajectory. Motivated by the failure modes we
observe in game development tasks, we ask whether an agent's exploration predicts the outcome,
and find that it does on both kinds of work. On tasks whose solutions have annotated
dependencies, and holding task and model fixed, inspecting the expected objects before acting on
them is associated with a $13.4$pp higher chance of passing on scene-only tasks ($95\%$ CI
$[+1.2,+25.7]$) and a $9.8$pp higher chance ($[+3.6,+16.0]$) on script-only tasks, whose work is
only to author script text (\S\ref{sec:grounded}).

The measurement scheme is not specific to game development. Any evaluation of tool-using
agents that labels its tools as observation or action can compute the label-based features, and
one that also annotates each task's dependencies can compute the coverage signal. We release all
task definitions, the place files they run against, the per-task annotations used in
\S\ref{sec:grounded}, a plugin that runs the tasks in \editor{}, and the leaderboard under the
MIT license at \repo{}.

\section{Related Work}
\label{sec:related}

Agent evaluation has become a broad subfield, with its own taxonomies of capabilities,
environments and benchmark dimensions \citep{yehudai2025survey}. We know of no prior work that
combines three elements: an agent working inside a deployed game engine, correctness judged by
how the running game behaves under simulated player input and unit tests, and the agent's
exploration instrumented as a measured signal. \S\ref{app:related} covers the related
literature on code generation and benchmark validity.

\textsc{WebArena}, \textsc{OSWorld} and \textsc{AgentBench}
\citep{zhou2023webarena,xie2024osworld,liu2023agentbench} evaluate agents in live, stateful
environments, and \textsc{GAIA} \citep{mialon2023gaia} evaluates assistants on real-world questions that need
web browsing and tool use. All of them report end-to-end success, and none is built around
writing software. \textsc{Terminal-Bench} \citep{terminalbench2026} is, with agents working inside
a stateful container, but by design its tests check only the final state, not the agent's
commands. None of these benchmarks annotates what a correct solution has to consult.

\textsc{AgentBoard} \citep{ma2024agentboard} also looks beyond binary success. Across its nine tasks it reports a progress rate that reaches one at completion. This rate is the running
maximum, over steps, of how close the agent's \emph{state} has come to the goal, so it is a
finer-grained measure of outcome. We instead measure a property of the investigation: the
fraction of a task's annotated dependencies the agent consulted.

\citet{jiang2026exploration} study exploration directly. They build partially observable grid
environments over an unknown task DAG, evaluate $13$ frontier models, and find that exploration error strongly predicts success across models ($R^2{=}0.95$ for log exploration error). Consistent with our results, they show that models
with indistinguishable success rates behave in qualitatively different ways. The level of
analysis differs: they report a correlation across models, while we relate exploration to
success across repeated runs with task and model held fixed.

\textsc{GameDevBench} \citep{chi2026gamedevbench} is the closest work to ours. Its verification
is also in-engine and deterministic, decided by Godot unit tests rather than by a judge. Its
verifiers inspect scene and runtime state. Ours drive a gameplay session with simulated player
input and assert on what the client observes, which requires a running server and a driven
client (\S\ref{app:setup}). Our focus on exploration and trajectories is the other main
difference. \textsc{GameCraft-Bench} \citep{luo2026gamecraft} instead scores the delivered game
with a multimodal judge (\S\ref{app:related}). 

On the process side, \citet{mehtiyev2026beyond} analyze $9{,}374$ trajectories from $19$ agents
on $500$ SWE-bench Verified tasks. After controlling for task difficulty, what remains predictive
is trajectory \emph{structure}: agents that gather context before editing and invest in
validation succeed more often. Their behavioral features are the same for every task. Ours are
scored against each task's own reference solution and annotated dependencies, so we can ask
whether the agent investigated the objects and scripts that a specific task requires, not just
how much it investigated in general.  

\section{The \toolname{} Framework}
\label{sec:framework}

\toolname{} is designed for \emph{fidelity} (physics, networking, and gameplay behave as a
creator or player would experience them), \emph{harness agnosticism} (any agentic system can be
evaluated on identical tasks without modifying the environment), and \emph{verifiability}
(machine-checkable assertions decide success). Each evaluation opens a real placefile, a base
game inside the editor, and runs a candidate agent inside it. The framework ships seventeen
curated \plat{} experiences, ranging from an empty baseplate to a racing game, a laser-tag arena,
a suburban town and a modern city. The $113$ scored tasks use eleven of them: $70$ tasks are set
in a populated game and $43$ start from a bare placefile ($42$ baseplates and one flat terrain).
Every task has a human-written reference solution and unit tests. The public repository includes
$67$ of the $113$ reference solutions, and the rest are withheld.

The agent works through eight tools, which we group into \emph{observation} and
\emph{action}. The engine calls every node of a place's object hierarchy (a part, a model, a
folder, a script, a service) an \emph{Instance}. The tool names and our annotations use this
term, so from here on we write ``instance'' for what the earlier sections called an object.

\begin{center}
\small
\begin{tabular}{@{}p{0.46\linewidth}p{0.46\linewidth}@{}}
\toprule
\textbf{Observation (reading hidden state)} & \textbf{Action (mutating state)} \\
\midrule
\code{search\_game\_tree} (traverse scene graph) & \code{multi\_edit} (edit script source) \\
\code{script\_read} (read a script) & \code{execute\_luau} (run engine code) \\
\code{script\_grep} (content search) & \code{insert\_from\_creator\_store} \\
\code{script\_search} (fuzzy name match) & \\
\code{inspect\_instance} (read properties) & \\
\bottomrule
\end{tabular}
\end{center}

Separating observation from action is the paper's main methodological choice. It lets us say
where two models with equal pass rates differ (in observation, in action, or in the mix between
them) instead of reporting a single tool-call count. \S\ref{app:setup} describes the task
lifecycle, the two-phase check structure, the authoring pipeline, and what can be reproduced
from the release.

Every result in this paper is a property of the model together with one fixed scaffold, which
we describe here (\S\ref{app:setup} gives details). Each attempt delivers the task prompt as a
single user turn. The model then runs its own tool-calling loop: tool calls are accepted
automatically, there is no follow-up prompting, and the whole conversation is carried forward
without truncation. A failed tool call is not retried. Its error message is returned to the
model as an ordinary tool result. The loop ends when the model stops calling tools, when the
$300$\,s wall-clock budget runs out, or when the run reaches $200$ tool calls, whichever comes
first. Almost every attempt ends the first way: $98.6\%$ of core attempts stop because the model
stopped calling tools, the remaining $1.4\%$ run out of time, and no run reaches the tool-call
ceiling. All thirteen models use the same system prompt, which is based on the Roblox assistant's prompt and cannot
be released. Sampling parameters are fixed per model family, not per model or per run, and
reasoning effort is set to \code{high} for every family that accepts a reasoning level. All
attempts were collected between May and July~$2026$.

Each task's checks run in two phases. \code{check\_scene} asserts properties of the edited
world, and \code{check\_game} asserts behavior in a play session started after the edits. A task
passes only if every phase passes (\S\ref{app:setup}).

\section{The \toolname{} Benchmark}
\label{sec:benchmark}

The $113$ scored tasks are what remains of $117$ after the check-validity screens of
\S\ref{sec:limitations}. They form two suites that we score separately: $84$ core tasks and $29$
debugging tasks. \S\ref{app:catalog} lists all $113$ with their prompts, place files, and check
phases. A core task gives the model a place and a feature request, and the model must decide
what to build and where. Most core tasks open on a populated game ($63$ of $84$). The other $21$
($20$ baseplate, $1$ flat terrain) start the model in an empty world, where the work is
construction rather than discovery. A debugging task gives the model a place with a planted
defect and asks for a repair, so the target is located and verifiable.

\begin{figure}[tbp]
\centering
\includegraphics[height=0.25\linewidth]{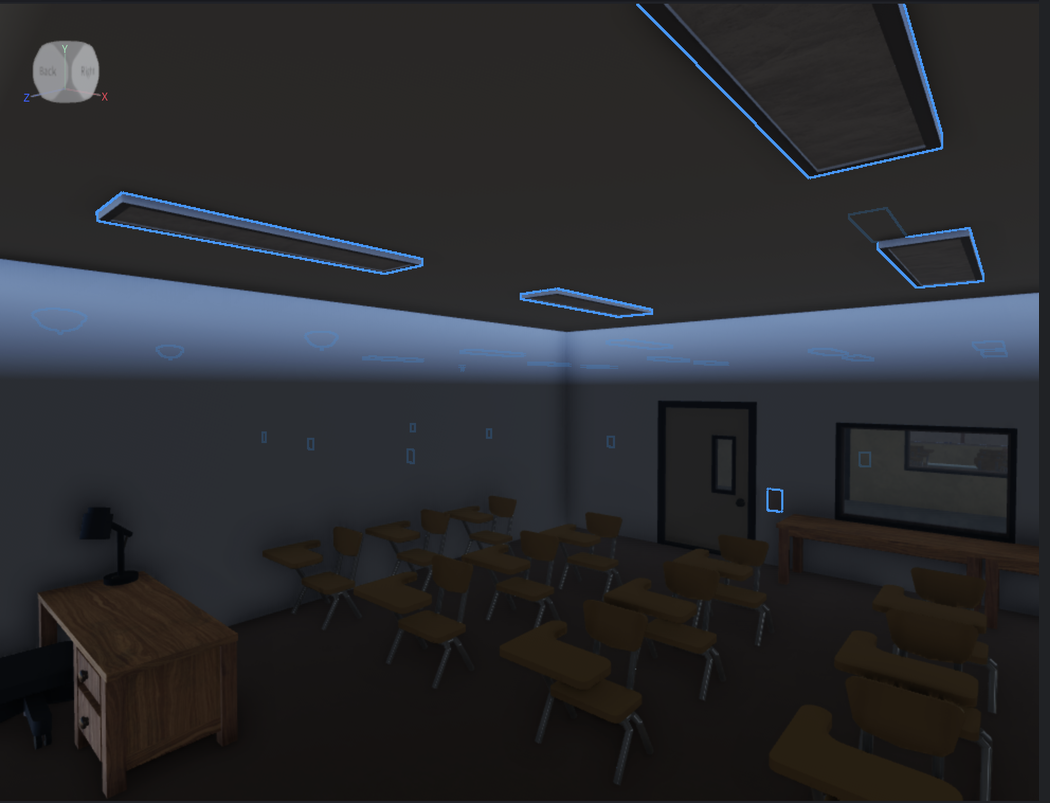}\hfill
\includegraphics[height=0.25\linewidth]{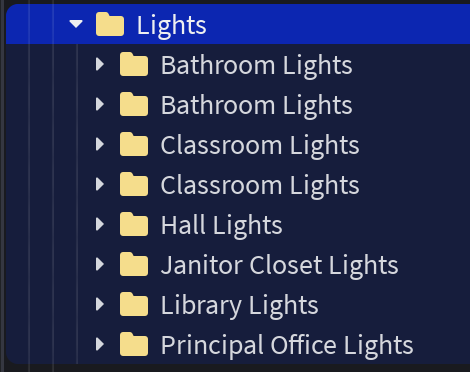}\hfill
\includegraphics[height=0.25\linewidth]{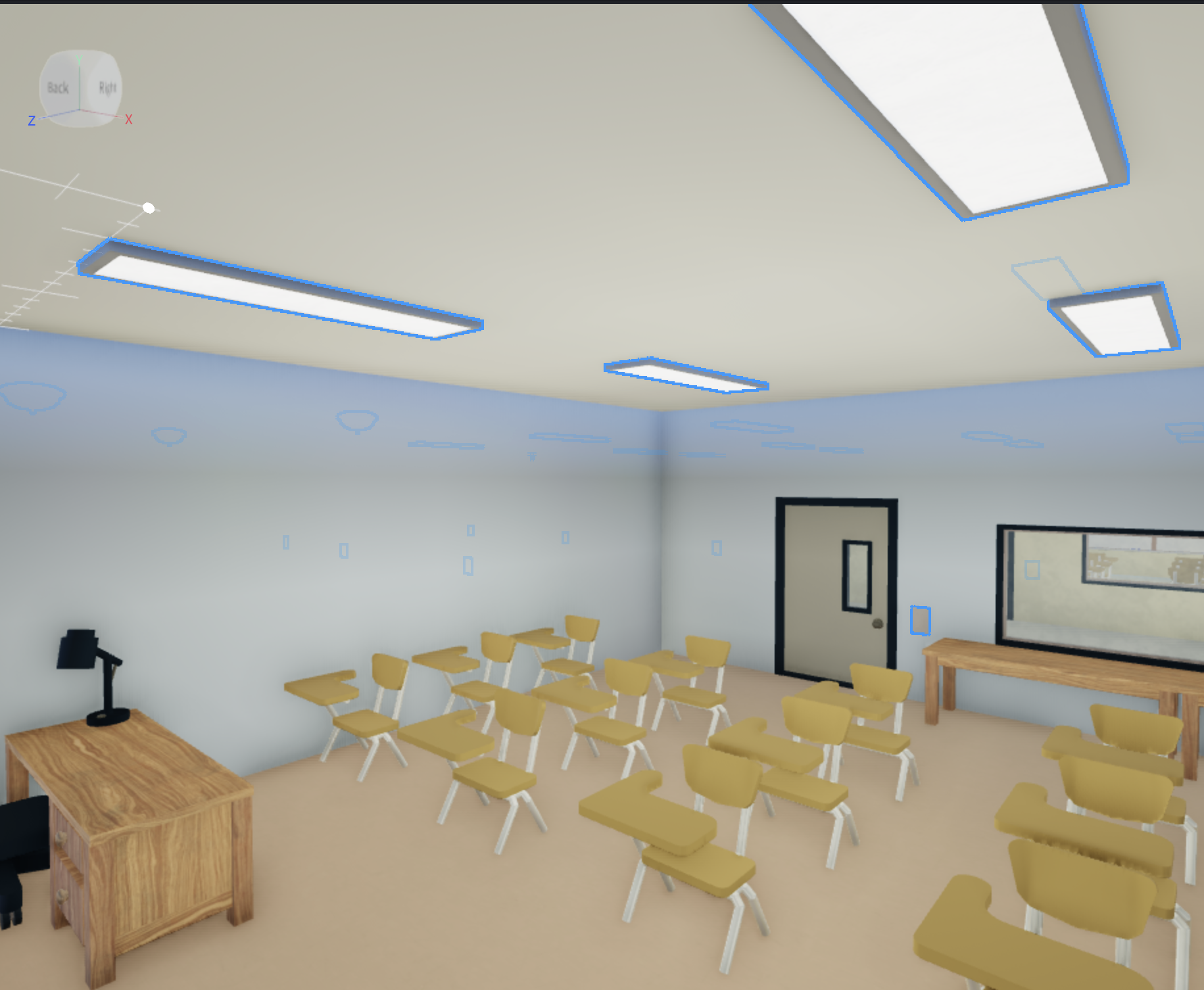}
\vspace{2pt}
\caption{\textbf{A worked core task: \code{080\_surburban\_school\_lights\_on}.} The entire
prompt is \emph{``Make every light in the school start in the `on' state.''} \emph{Left:} the
scene as the agent receives it. \emph{Center:} the editor's object explorer at \code{Workspace/School/Lights}. The prompt
names none of the eight per-area subfolders, the \code{LightsOn} boolean each carries, or the
switch scripts that read it. \emph{Right:} the
classroom with the lights on. The check reads the resulting instance state (\S\ref{app:setup}).
Pass@1 spans $0\%$ to $100\%$ across the $13$ models.}
\label{fig:env}
\end{figure}

The core suite is our primary object of analysis. Every estimate, interval and test is
computed within one suite and never pooled across the two. The debugging suite appears next to
the core partitions in Figure~\ref{fig:kinds}, and Table~\ref{tab:debug}
(\S\ref{app:leaderboard}) reports it per model. No model solves six of the 84 core tasks in any
of their 16 attempts. All six are solvable: each task's own reference solution passes its
checks, so no check is impossible to satisfy. We also read each of the six checks against its
prompt and found no mismatch. Four more tasks are solved on every attempt. Together, the two
groups shrink the usable core suite to the $74$ tasks that carry any between-model information
(\S\ref{app:irt} names all ten). Every estimate still runs over all $84$: the ten change no
model's ranking and add nothing to a paired comparison. Figure~\ref{fig:env} walks through one
core task from start to finish.

\paragraph{Two kinds of work: script authoring and scene change.}
We also divide the core suite by the kind of work each task's reference solution requires,
using our annotations. A task is \emph{script-authoring} if its solution requires
\code{multi\_edit}, meaning it must write or rewrite script text. Every other task is
\emph{scene-change}: it is completed without writing script text, by running \lang{} code that
mutates the live game scene. This rule labels every core task and splits the suite $59$ to
$25$. The two classes are a requirement and its complement, not two disjoint activities. Of the
$59$ script-authoring tasks, $29$ also require \code{execute\_luau}. All $25$ scene-change tasks
require \code{execute\_luau}, and none requires \code{multi\_edit}. \S\ref{sec:grounded} uses a
stricter, non-overlapping split. It sets aside the $29$ tasks that require both tools and scores
the $30$ \emph{script-only} and $25$ \emph{scene-only} tasks that remain.

\section{Metrics}
\label{sec:metrics}

\paragraph{Outcome metrics.}
For each task we draw $k{=}16$ independent attempts and estimate four success criteria at a
subsample size of $5$. Throughout, $k$ is the number of attempts per task, and the number after
the @ is the subsample size a criterion is named for. Instead of plugging in raw sample rates, we
use unbiased U-statistic estimators. pass@1 is the per-task success rate. pass@5, cons@5 and
all@5 are the probabilities that a uniformly random subsample of $5$ of a task's $16$ attempts
contains at least one, at least three, and five successes. They are computed in closed form from
the $16$ draws, not by extrapolation. These estimators follow the unbiased pass@$k$ estimator of
\citet{chen2021codex}, extended from ``at least one success'' to the consensus and unanimity
criteria. The spread between pass@5 and all@5 is a consistency gap. Our all@5 is the pass$^k$
reliability metric of $\tau$-bench \citep{yao2024taubench}, and Figure~\ref{fig:reliability}
(\S\ref{app:leaderboard}) traces it.

\paragraph{Process metrics.}
We report per-tool usage, split into observation and action, and a per-call tool-error rate.
We define exploration as how much of what a task depends on the agent reaches through its
observation tools before acting on it. Observation that only verifies an edit already made
therefore does not count as investigation (\S\ref{sec:grounded}).

\paragraph{Inference.}
Every comparison between two models is paired. We take the mean per-task difference in pass@1
over the matched core tasks and compute a percentile bootstrap interval by resampling tasks. A
two-sided Wilcoxon signed-rank test on the same per-task differences serves as a check. A pair
counts as resolved when the bootstrap interval excludes zero. To decide whether an individual
task flips between two models, we apply Fisher's exact test to that task's $2\times2$ table of
passes and failures for the two models over their $16$ attempts each. Pairing is what makes a
suite of this size sensitive: it removes the per-task difficulty that both models face, which
dominates the spread across tasks. \S\ref{app:stats} gives the resampling settings, the suite's
resolution, and further details.

\section{Core-Suite Results}
\label{sec:frontier}

\begin{table}[tbp]
\centering
\caption{\footnotesize\textbf{Core-suite results for all 13 models}, each running the same
$84$ core tasks at $k{=}16$ attempts under identical system prompts and tools, with a
$300$\,s timeout counted as a failure.
``Tools'', ``Obs'' and ``Obs\%'' are mean tool calls per run, mean observation calls per run,
and their ratio, from the transcripts. ``Err'' is errors per tool call, and ``Tmo'' is the share
of a model's attempts that exhaust the $300$\,s time budget. Pass-rate subscripts are marginal
standard errors from the task bootstrap of \S\ref{app:stats}, which describe how precisely each
model's own suite-level rate is estimated. Every model-versus-model claim in the paper rests on
paired comparisons. Best per column in \textbf{bold}. \S\ref{app:leaderboard} gives the
reliability ladder and the debugging suite.}
\label{tab:leaderboard}
\footnotesize
\setlength{\tabcolsep}{4pt}
\begin{tabular}{@{}lccccccccc@{}}
\toprule
Model & Pass@1 & Pass@5 & Cons@5 & All@5 & Tools & Obs & Obs\% & Err (\%) & Tmo (\%) \\
\midrule
Claude Fable 5         & $\mathbf{51.7}_{(4.9)}$ & $63.7_{(4.9)}$ & $51.8_{(5.3)}$ & $\mathbf{39.4}_{(5.1)}$ & 6.4 & 3.9 & 62 & 1.17 & 2.31 \\
Gemini 3 Flash         & $51.6_{(4.7)}$ & $\mathbf{66.7}_{(4.9)}$ & $51.8_{(5.2)}$ & $34.5_{(4.5)}$ & 10.8 & 7.0 & 65 & 1.94 & 0.15 \\
Claude Opus 4.8        & $51.5_{(4.7)}$ & $65.9_{(5.0)}$ & $\mathbf{52.4}_{(5.1)}$ & $35.0_{(5.0)}$ & 7.0 & 4.8 & 68 & 1.31 & 2.16 \\
Gemini 3.1 Pro         & $51.3_{(4.7)}$ & $66.6_{(4.8)}$ & $50.9_{(5.2)}$ & $36.7_{(5.0)}$ & 7.8 & 4.1 & 52 & 2.95 & \textbf{0.00} \\
Gemini 3.5 Flash       & $50.6_{(4.7)}$ & $66.1_{(4.9)}$ & $51.3_{(5.2)}$ & $33.7_{(4.8)}$ & 12.9 & 8.3 & 64 & 2.12 & 1.64 \\
GPT-5.6-sol            & $47.3_{(4.9)}$ & $60.5_{(5.2)}$ & $46.5_{(5.2)}$ & $35.4_{(5.1)}$ & 10.4 & 8.5 & 82 & 0.89 & 2.31 \\
GPT-5.5                & $46.1_{(4.7)}$ & $61.6_{(5.0)}$ & $46.3_{(5.1)}$ & $30.5_{(4.9)}$ & 8.7 & 6.9 & 79 & \textbf{0.74} & 1.49 \\
Claude Opus 4.7        & $45.5_{(4.7)}$ & $60.8_{(5.1)}$ & $45.4_{(5.1)}$ & $30.7_{(4.8)}$ & 5.0 & 3.0 & 59 & 2.82 & 0.82 \\
GLM-5                  & $42.9_{(4.4)}$ & $61.2_{(5.1)}$ & $45.3_{(5.1)}$ & $20.3_{(3.8)}$ & 8.4 & 5.9 & 70 & 2.48 & 1.04 \\
GPT-5.6-terra          & $40.4_{(4.8)}$ & $53.5_{(5.2)}$ & $40.3_{(5.2)}$ & $27.5_{(4.7)}$ & 9.7 & 7.6 & 78 & 1.05 & 1.56 \\
GPT-5.6-luna           & $40.3_{(4.6)}$ & $56.2_{(5.1)}$ & $40.5_{(5.1)}$ & $24.9_{(4.6)}$ & 12.0 & 9.7 & 81 & 1.44 & 0.74 \\
GPT-5.4                & $40.0_{(4.7)}$ & $55.0_{(5.2)}$ & $40.3_{(5.2)}$ & $24.5_{(4.4)}$ & 12.1 & 10.3 & 85 & 1.14 & 3.57 \\
GLM-4.7                & $36.7_{(4.3)}$ & $56.0_{(5.0)}$ & $37.3_{(4.9)}$ & $17.0_{(3.8)}$ & 8.6 & 5.6 & 64 & 3.74 & \textbf{0.00} \\
\bottomrule
\end{tabular}

\end{table}

The five leading models are within $1.1$pp of one another, between $50.6$ and $51.7\%$
(Table~\ref{tab:leaderboard}). Their $95\%$ marginal intervals overlap, and so do those of every
other model pair. Overlapping marginal intervals are not evidence that two models are
equivalent, so we compare models on their paired per-task differences rather than on separate
summary statistics \citep{miller2024errorbars}. A paired test on the shared tasks separates $34$
of the $78$ pairs, and $15$ of them survive a Holm correction.

Similar average pass rates do not mean these models behave similarly, and the sets of tasks
they solve show this directly. Counting a task as solved when a model passes a majority of its
$16$ attempts, each of the top five models solves between $41$ and $44$ of the $84$ core tasks.
If they all solved the same half of the suite, the union of their solved sets would be about that
size too. Instead, the union contains $58$ tasks and the intersection $27$. So $31$ tasks, over a
third of the suite, are solved by at least one of these models and missed by at least one
other.

These agreement numbers should not be compared against a baseline of zero. ``Solved'' is a
majority verdict on $16$ noisy attempts, so two identical models can land on opposite sides of
the threshold whenever a task's true success rate is near one half. A union larger than the
intersection is therefore expected even when nothing distinguishes the models. We estimate the
right baseline with a pooled placebo. For each task, we pool the five models' successes and
attempts into a single success rate. This keeps each task's difficulty but erases every
difference between the models. We then redraw all five models from that common rate, $16$
attempts each, apply the same majority threshold, and record the union and intersection. Over
$2{,}000$ draws, the union has a median of $47$ tasks ($95\%$ interval $[45,50]$) and the
intersection a median of $37$ ($[34,40]$). The observed $58$ and $27$ fall outside both
intervals, and none of the $2{,}000$ draws reaches either value. The divergence is not an
artifact of thresholding sixteen attempts. 

Comparing models pair by pair gives a sharper picture. For each of the ten pairs among the top
five models, we compare the two models' $16$-attempt success counts on each task with Fisher's
exact test and count the tasks that separate at $p<0.05$. The median across the ten pairs is
$17$ tasks, with a range of $9$ to $21$. We set a floor with the same kind of placebo. For each
task we pool the pair's $32$ attempts into a single rate, redraw both models from that rate so
that they differ only by chance, and run the same test on the simulated counts, again over
$2{,}000$ draws. The placebo flags $0.6$ tasks on average, and no pair's $95$th percentile
exceeds $2$. 

Across all $78$ pairs in the panel, the median number of tasks separating at $p<0.05$ is $19$.
Two models at the top of the leaderboard, a tenth of a point apart in aggregate, therefore
disagree about individual tasks nearly as much as two models ten points apart. The disagreements
run in both directions in nearly equal numbers, which is why they cancel in the average.
Claude~Fable~5 and Gemini~3~Flash differ by $0.1$pp in aggregate but on $17$ individual tasks,
$8$ favoring one model and $9$ the other. 

\begin{figure}[tbp]
\centering
\includegraphics[width=0.98\linewidth]{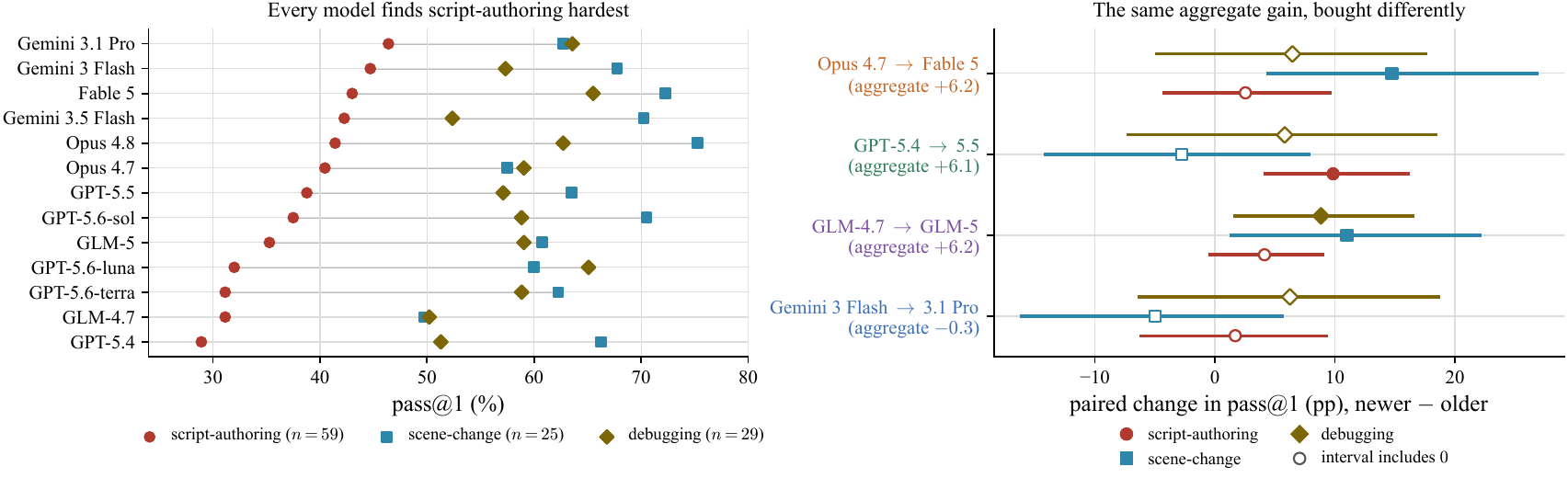}
\caption{\textbf{One pass rate hides what a model is actually good at.} Core tasks are split by what their solution must do: author or rewrite script text ($59$
tasks) or mutate the game scene ($25$). The debugging suite ($29$) is scored separately.
\emph{Left:} pass@1 per group, with models ordered by script-authoring rate. \emph{Right:} four
within-family generational steps, broken down by group, with $95\%$ paired-bootstrap intervals
over tasks. Filled markers indicate intervals that exclude zero.}
\label{fig:kinds}
\end{figure}

Splitting by kind of work shows what the models disagree about. For the same five models, the
spread between the best and the worst grows from $1.1$pp on the full core suite to $5.0$pp on
script-authoring tasks, $12.5$pp on scene-change tasks, and $13.1$pp on the debugging suite.
Under the same placebo, five identical models give median spreads of $2.3$pp and $4.0$pp on the
two kinds of work. The observed spreads are therefore $2.2$ and $3.1$ times their floors, while
on the full suite the observed $1.1$pp is below its floor of $2.0$pp. The rankings also change
(Figure~\ref{fig:kinds}, left). Gemini~3.1~Pro is first of thirteen at script authoring and
eighth at scene change. Claude~Opus~4.8 is the reverse: first at scene change and fifth at script
authoring. What looks like a single frontier at $50\%$ is five differently shaped capability
profiles that average to the same number. The five also use tools in visibly different ways:
mean tool calls per run range from $6.4$ to $12.9$, and the observation share of those calls
from $52\%$ to $68\%$ (Table~\ref{tab:leaderboard}). 

The same pattern holds across model generations (Figure~\ref{fig:kinds}, right). Claude, GPT
and GLM each gain about $6$pp in core pass@1 from their older to their newer model, but on
different kinds of work. Gemini is flat in aggregate. 

To ask which tasks separate models and what those tasks have in common, we fit a 2PL
item-response model to the $13\times84$ response matrix. With only $13$ respondents, the fit is
a diagnostic rather than a precise measurement. \S\ref{app:irt} gives the specification, priors
and diagnostics. When ability is modeled hierarchically by provider family, $79\%$ of its
variance lies between families rather than between versions within a family (interval
$[0.71,0.87]$).

Seven tasks have a discrimination $\alpha$ whose $95\%$ posterior interval lies entirely above
the suite's median under both priors we fit (\S\ref{app:items}). They are turning lights on and
off by time of day ($14\%$ pass rate), making the player bounce when shooting the ground within
$5$ studs ($19\%$), a gravity well ($82\%$), a flashlight for every joining player ($83\%$),
friendly fire in a laser-tag game ($79\%$), a fridge door that opens on touch instead of click
($67\%$), and a pasted error about a missing \texttt{getDriversInPart} member ($90\%$). Among the
$22$ mid-difficulty tasks ($|\beta|\le0.5$), discrimination is essentially unrelated to pass rate
(Spearman $0.12$), so a task cannot be made discriminating by aiming it at a $50\%$ pass rate.
What they share instead is an expectation the prompt implies but does not state. For instance,
the lighting task needs a day--night cycle, the bounce must not fire at walls, ceilings or the
sky, friendly fire must reach every blaster, and a flashlight must carry an actual light. Fixing
the \code{getDriversInPart} error means writing a function that actually returns the drivers in a
race, not one that merely makes the error go away.

\section{Exploration Behavior as a Predictor of Task Success}
\label{sec:grounded}

Much of the difference between models lies in \emph{which} tasks they solve. We now ask
whether the \emph{trajectory} of a run carries information about whether that run succeeds. For
each trajectory on the core panel, we measure whether the run carried out the investigation the
task calls for. We use the per-task dependencies annotated in the benchmark: the specific
instances the reference solution touches and the specific scripts it reads.
Figure~\ref{fig:traj} in \S\ref{app:traj} shows representative runs. 

We regress the $0/1$ outcome of a run on one trajectory feature at a time by least squares,
with a dummy for every task and every model. Confidence intervals use a cluster-robust sandwich
estimator with tasks as clusters. The fixed effects make every estimate \emph{within task and
within model}: we compare runs of the same model on the same task that went about it
differently. Because the outcome is an indicator, each coefficient is directly in percentage
points.

A reference-solution instance counts as ``observed'' only when the run reaches it through an
\emph{observation} tool. The observation must also \emph{precede} the run's first action on that
instance. A run that edits the right object and only then inspects it to confirm the edit gets
no credit for investigating before acting. A run's \emph{coverage} on a task is the fraction of
the task's annotated dependencies credited this way. It is $0$ for a run that acts without
inspecting any of them and $1$ for a run that reaches all of them before acting. Observing an
instance is not the same as modifying it correctly, so coverage does not simply restate the
checks that decide success. Table~\ref{tab:grounded} reports the estimates. Coverage is defined
only where a task's dependencies are annotated, and $54$ of the $84$ core tasks carry the
annotation. We analyze $37$ of them, split into $24$ \emph{scene-only} and $13$
\emph{script-only} tasks. The remaining $17$ require both tools and are reported in
\S\ref{app:traj}.

On the \textbf{scene-only} tasks ($24$ tasks, $13$ models), the bottleneck is finding and
acting on the right live objects in the game. There, each standard deviation of coverage
predicts a $+6.4$pp higher pass probability ($+13.4$pp from none to full, $95\%$ CI
$[+1.2,+25.7]$). The \textbf{script-only} tasks ($13$ tasks, $13$ models) require authoring
script text and no scene mutation. On them, coverage shows an association of similar size:
$+4.3$pp per standard deviation and $+9.8$pp from none to full (CI $[+3.6,+16.0]$).

\begin{table}[tbp]
\centering
\caption{\footnotesize\textbf{The grounded success signal is present on both kinds of work, at similar magnitudes.}
Effects on the probability that a run passes, in percentage points, from a linear probability
model with task and model fixed effects over the per-run trajectories. Coverage features count a
reference-solution dependency only when the run reaches it through an observation tool
\emph{before} its first action on that dependency. Expected-tool match is the fraction of the
observation tools a task is annotated as requiring that the run used at least once, in any
order. ``/SD'' is the effect per standard deviation of the feature, computed within the arm the
column reports and over the tasks that row uses. ``none$\to$full'' is the same coefficient,
scaled to a change from none to full. Brackets are $95\%$ task-cluster-robust intervals. The
difference column is the coefficient on the feature's interaction with a script-arm indicator,
from one fit over both arms, which gives the gap its own cluster-robust interval. It need not
equal the difference between the two columns, because each per-SD figure is scaled by its own
arm's feature dispersion. The lower block shows two sequencing features that use no annotated
dependency information. \S\ref{app:traj} gives cluster counts per row, the feature definitions,
and alternative inference schemes.}
\label{tab:grounded}
\scriptsize
\setlength{\tabcolsep}{3pt}
\begin{tabular}{@{}lccc@{}}
\toprule
Trajectory feature & Scene-only & Script-only & Difference \\
\midrule
Observe reference instances (/SD) & $+6.4[+0.6,+12.3]$  & $+4.3[+1.6,+7.0]$   & $-1.6[-7.6,+4.3]$   \\
\quad\quad none$\to$full          & $+13.4[+1.2,+25.7]$ & $+9.8[+3.6,+16.0]$  & $-3.5[-16.3,+9.3]$  \\
Read reference scripts (/SD)      & --                  & $+8.3[+1.1,+15.5]$  & --                  \\
\quad\quad none$\to$full          & --                  & $+17.8[+2.3,+33.3]$ & --                  \\
Expected-tool match (/SD)         & $+5.3[-2.5,+13.0]$  & $+0.5[-3.4,+4.4]$   & $-6.8[-17.8,+4.1]$  \\
\midrule
Search, then read it (/SD)        & $+8.6[-3.3,+20.5]$  & $-0.3[-4.3,+3.7]$   & $-8.9[-21.7,+3.9]$  \\
Reads that were grounded (/SD)    & $+4.0[+0.4,+7.6]$   & $+1.6[-2.1,+5.3]$   & $+1.4[-3.5,+6.4]$   \\
\bottomrule
\end{tabular}
\end{table}

Scripts are a second investigation channel. On the $11$ tasks whose reference solution
depends on other scripts ($7$ of them outside the two arms, \S\ref{app:traj}), reading those
scripts before making edits predicts a $+8.3$pp higher pass probability per standard deviation
($+17.8$pp from none to full, $[+2.3,+33.3]$). On script work, then, the signal also runs
through the scripts a task depends on. We cannot test whether scene work has a script-side
counterpart, because only $2$ scene tasks carry that annotation.

The other rows of Table~\ref{tab:grounded} do not credit the specific instances a reference
solution touches. Expected-tool match rewards using the annotated observation tools in any
order, and the two sequencing features in the lower block use no annotation at all
(\S\ref{app:traj}). None of them shows an effect on the script-only arm, and on the scene-only
arm only grounded reads is significant. This is consistent with the signal lying in which
instances a run reaches, not in which tools it calls.

The two arms are estimated on disjoint task sets, so comparing their estimates side by side
does not show whether they differ. We estimate the gap directly, with one fit over both arms
that includes coverage and its interaction with a script-arm indicator. The gap is small and not
significant: $-1.6$pp per standard deviation, cluster-robust interval $[-7.6,+4.3]$, $p{=}0.59$
across the $37$ task clusters. We detect no difference between the two kinds of work, and
coverage predicts success on both at magnitudes we cannot tell apart. If observation is credited
at any point in the run, rather than only before the first action on the dependency, the scene
effect is about twice the script effect. That ratio comes from the unrestricted feature. The
timing restriction lowers the scene estimate from $+7.8$ to $+6.4$pp per standard deviation and
does not affect the script estimate.

The amount of exploration does not explain the effect. Controlling for the number of tool
calls moves the scene estimate only from $+13.4$ to $+12.6$pp, even though the number of calls
predicts success on its own on that arm ($+5.6$pp per standard deviation). A sharper test uses a
feature that measures context-gathering \emph{without} knowing what the task depends on, which
is the form the finding of \citet{mehtiyev2026beyond} takes on SWE-bench. We measure it as the
number of distinct instances a run inspects before its first action. On the same runs and
tasks, this generic feature predicts $+6.2$pp per standard deviation on the scene arm
($[+1.9,+10.5]$), almost as much as coverage, and $+2.7$pp on the script-only arm
($[-2.4,+7.9]$), where it is not significant.

To test whether coverage measures anything the generic feature does not, we enter both in the
same fit. On scene-only tasks the two features correlate at $+0.55$, and neither stands out:
with both in the model, coverage takes $+5.0$pp $[-1.1,+11.1]$ and the generic feature $+4.1$pp
$[-0.4,+8.6]$. On script-only tasks they come apart. They correlate at only $+0.32$, coverage
holds at $+3.9$pp $[+0.8,+6.9]$, and the generic feature drops to $+1.2$pp $[-4.6,+7.0]$. On
script-only work, knowing \emph{which} scripts and instances a task depends on carries
information that a plain count of exploration does not. On scene-only work we cannot separate
the two.

Task and model fixed effects enter our model additively. Part of the relationship could
therefore come from differences \emph{between} (task,\,model) cells rather than within them: a
model that happened to do well on a particular task may also have happened to observe more
there, and additive effects do not absorb that. A stricter specification gives every
(task,\,model) cell its own fixed effect, leaving only the variation among one model's $16$
attempts at one task. Coverage survives it on both arms, at $+5.7$pp per standard deviation on
the scene arm ($[+0.2,+11.3]$) and $+2.8$pp on the script arm ($[+0.8,+4.8]$).

\paragraph{Implications for process supervision.}
Process reward models supervise reasoning at the step level \citep{lightman2024verify}, and
recent work scales automated process verifiers by asking which intermediate signals predict
eventual success \citep{setlur2024rewarding}. A verifier that rewards coverage of the instances
a task depends on, credited only when the agent reaches them before acting, would reward a
behavior worth $+9.8$pp of pass probability on script-only work and $+13.4$pp on scene-only
work. On script-only tasks, such a verifier would need to be task-specific: there the signal
survives the generic pre-edit exploration feature, so a verifier that rewards only how much an
agent explored would miss it. Our results also show that one can measure which process signals
carry information about success before training any verifier. These are within-task,
within-model associations, so they identify signals worth testing as process rewards, not
signals proven to work.

\section{Limitations and Threats to Validity}
\label{sec:limitations}

\toolname{} covers a single engine and a single language, so the trajectory signal we find
may not transfer unchanged to other engines. We conjecture that the link between task-required
investigation and success holds generally in stateful worlds with hidden structure, but we
cannot test this here. Our exploration measure is also a behavioral proxy tied to our eight-tool
action space. It captures how much a model probes the world through the provided tools, not what
the model may infer without a tool call. Coverage also depends on annotating each task's
dependencies from its reference solution, and none of the annotation-free alternatives we test
shows a significant association with success on script-only work. Deriving dependencies
automatically, for example from the instances and scripts a reference solution reads and writes
when executed, would remove the annotation step.

Agents in our setting observe the world only through text. Tool outputs describe instances,
their properties and script contents, and no tool returns a rendered frame. Visual feedback
matters for game development: adding image and video feedback raises model performance on
\textsc{GameDevBench} \citep{chi2026gamedevbench}. Our results therefore describe text-only
agents, and we do not test what vision would add on tasks judged by layout or visible
behavior.

Our checks could be wrong in two directions, and we screen for both. Running each task's
reference solution shows that a check is not too strict. A null arm that makes no edits at all
shows whether a check can be passed without doing the work. A third screen reads the definition
of every check that no model passes. Four tasks failed these screens and were dropped
(\S\ref{app:setup} lists them and the reasons). The null arm also passes $5$ of the $113$ scored
tasks, four core and one debugging. We keep these: their setup is intact, but their checks also
accept an untouched world, so they test whether a model preserves functionality that already
works. Three of the four core tasks still separate models, and the fourth separates none.

Our play-time checks are scripted rather than model-driven. This makes them deterministic, but
far from representative of real player behavior. Model-driven user simulators can approximate
real users when conditioned on detailed user profiles \citep{dou2025simulatorarena}, though a
measurable distributional gap to real users remains \citep{mehri2026gap}.

The tasks, checks, worlds and annotations are public, and a released editor plugin re-runs each
task's setup, reference solution and checks. The scaffold around them is specified closely
enough to reimplement (\S\ref{sec:framework}). However, the system prompt is part of a live
product and cannot be released, and the runtime that executes sessions is hosted rather than
open source.

Public release creates a risk of contamination. To test whether this risk grows across model
generations, we track the pass@1 gap between two cohorts of core tasks, the $46$ October tasks
whose solutions are withheld and the $38$ April tasks whose solutions are public, and measure
how that gap changes from older to newer models. We average these changes within each of four
model families. The comparison assumes that, without contamination, capability gains would
affect both cohorts similarly. The family effects split evenly in sign, with a mean of $-1.0$pp
and a $95\%$ interval of $[-6.9,+5.0]$, so we find no consistent evidence that newer models
improved disproportionately on either cohort (\S\ref{app:setup} gives the calculation). The test
cannot detect contamination that both generations share.

\section{Conclusion}
\label{sec:conclusion}

We introduced \toolname{}, a play-time-verified benchmark for agentic software development
inside a deployed game engine. The strongest of $13$ frontier models fails $48\%$ of the core
suite, its five leaders reach nearly identical aggregates on different tasks, and investigating
what a task requires before acting predicts whether a run succeeds. We release the suite, its
annotations and a leaderboard as a testbed for studying how agents act in stateful worlds.

\bibliographystyle{tmlr}
\bibliography{references}

\clearpage
\appendix

\section{Framework and Benchmark Details}
\label{app:setup}

\paragraph{Lifecycle and verification.}
A task file declares a natural-language prompt, a placefile, and four hooks. \code{setup}
changes the world and selection state before the agent runs. \code{reference} is an
authoritative solution, which the model never sees during evaluation. \code{check\_scene}
asserts edit-mode (static) properties, and \code{check\_game} (or client checks) asserts
play-mode (runtime) behavior. \S\ref{app:catalog} lists every scored task with its prompt, its
place file, and which of the two check phases it uses. Separating the two phases separates static
correctness from runtime behavior. For example, the health-regeneration task removes the
existing health script and installs a server-authoritative damage event. At play time, the check
fires that event and asserts that the health the client sees holds at $50$ for $1.4$\,s, reaches
exactly $60$ at $2.1$\,s, and reaches $70$ at $3.1$\,s. No static check could capture this
timing-dependent, server-to-client property.

The check for the worked task of Figure~\ref{fig:env} asserts that every \code{LightsOn}
boolean is true, every spotlight is enabled, and every light part has been changed from
\code{SmoothPlastic} to \code{Neon}. 

\paragraph{Agent scaffold.}
Each attempt consists of a single user message. Across all sampled runs, every user-role
message after the first is a tool result, so nothing prompts, clarifies or re-plans the agent
mid-run. Tool errors are returned to the model rather than retried. Of all tool calls, $1.6\%$
return an error, and $9.7\%$ of runs contain at least one error. The $200$ tool-call ceiling
never binds: the median core attempt makes $7$ calls, the $99.9$th percentile makes $52$, and the
largest attempt on the core panel makes $63$.

The time budget, tool-call budget, attempt counts, system prompt and tool configuration are
identical across all runs. The only settings that differ are the served model identifier, the
API family it is called through, and a worker count that affects throughput but not the agent.
Sampling is set per family, and Table~\ref{tab:sampling} records every request parameter.
Reasoning effort was set explicitly, not left to provider defaults, and at the same nominal
level, \code{high}, for all four families. An attempt that returned no scoreable result because
of an infrastructure failure, such as a rate-limit response or an upstream server error, was
relaunched. A timeout is not an infrastructure failure: it is scored as a failure and was never
relaunched. No attempt was relaunched because of its output.

\begin{table}[tbp]
\centering
\caption{\textbf{Request parameters, by family.} The sampling parameters used by the harness.
A dash is a parameter we did not send. Reasoning effort was \code{high} for all four
families, under each provider's own parameter name.}
\label{tab:sampling}
\small
\setlength{\tabcolsep}{6pt}
\begin{tabular}{@{}llllll@{}}
\toprule
Family & Temperature & top-$p$ & top-$k$ & min-$p$ & Output cap \\
\midrule
Claude (3 models) & --- & --- & --- & --- & $64{,}000$ \\
Gemini (3 models) & $1.0$ & $0.95$ & --- & --- & $32{,}768$ \\
GPT (5 models) & --- & --- & --- & --- & --- \\
GLM (2 models) & $1.0$ & $0.95$ & $20$ & $0$ & --- \\
\bottomrule
\end{tabular}
\end{table}

The eight tools of \S\ref{sec:framework} are a subset of the tools the engine's assistant
exposes, which the engine vendor documents publicly (\mcpdocs{}). The full set is larger. It
also includes asset and material generation, playtest control, simulated input, documentation
retrieval and subagent delegation, and some tool names there have changed since our transcripts
were recorded. Only the eight listed tools were enabled for the runs reported here, so
generation, retrieval and subagent delegation were all disabled.

\paragraph{Reproducibility.}
The engine version is pinned, and checks are written to tolerate benign non-determinism. A
fixed $300$\,s per-phase wall-clock budget is held constant across all models and treated as part
of the quality bar. The framework simulates player interaction programmatically (button clicks,
keyboard input, camera manipulation), so tasks that require user action can be verified at play
time. The task definitions, their place files, and the per-task annotations that
\S\ref{sec:grounded} uses are available in \repo{}. The release also includes an editor plugin
that runs a task's setup, reference solution and checks, first in edit mode and then in play
mode, so anyone can repeat the screen for over-strict checks. The plugin does not invoke an
assistant. The service that drives a model's session inside \editor{} is not open source, so
the released artifacts fix \emph{what} is evaluated but not \emph{how} it is run
(\S\ref{sec:limitations}).

\paragraph{Construction.}
Tasks are built with input from domain experts and game developers. For each task we build or
select a placefile that provides context, and we hand-write the executable checks and an
authoritative reference solution. Every scenario was reviewed by people for comprehensiveness,
generalizability and stability. The suite is public and grew through dated releases, each
recorded in the repository's release log. The repository was created on October~13, $2025$, and
the first $47$ core tasks were released the next day. Another $30$ debugging tasks followed on
March~3, $2026$, and a further $40$ core tasks on April~14, $2026$ brought the core suite to
$87$. The tasks have therefore been public for different lengths of time, so their exposure to
any training corpus is uneven across the suite.

The repository held $117$ task files. We collected runs on all of them and then dropped four
after data-quality checks: one from the October core release, two from the April core release,
and one from the debugging suite. Three of these have since had their checks repaired in the
public repository. The check of \code{001\_make\_cars\_faster\_bug\_1} crashed on a new version
of the engine. The check of \code{098\_pirate\_lose\_health\_underwater} relied on a
precondition the harness never established. The check of \code{120\_food\_spawn\_hunger\_bar}
demanded behavior that the prompt did not make clear. The fourth task,
\code{073\_homestore\_dynamic\_pricing}, is passed by the null arm (\S\ref{sec:limitations}) on
every attempt. Its setup never planted the hardcoded price that the prompt asks the model to fix,
so as run, the task had nothing to repair. We withdrew it while we work on a better replacement.
This leaves the $113$ tasks scored in this paper, $84$ core and $29$ debugging.

\paragraph{Exposure and contamination.}
Of the $84$ core tasks, $46$ come from the October $2025$ release and have no public reference
solution, and $38$ come from the April $2026$ expansion and have one. The two cohorts differ in
difficulty, so we do not interpret their raw pass-rate difference. Instead, for each model we
compute the mean task-level pass@1 on each cohort and subtract the published-cohort mean from
the withheld-cohort mean. For each of the nine older-to-newer model pairs in our four families,
we subtract the older model's cohort gap from the newer model's, and we average these changes
within each family. The family-level changes are $+0.7$pp for Claude, $+2.9$pp for GPT,
$-1.6$pp for GLM, and $-5.8$pp for Gemini. Their mean is $-1.0$pp, with a $95\%$ Student-$t$
interval of $[-6.9,+5.0]$ that treats the four families as independent units.

A positive change means that newer models gained more on the October cohort, whose prompts and
checks had been public longer. A negative change means that they gained more on the April
cohort, whose reference solutions are public. The interval is wide and includes zero, so this
test shows no differential improvement between the cohorts.

\paragraph{Null-arm validation.}
The null arm of \S\ref{sec:limitations} is a do-nothing agent: the task's checks run against the
world its setup leaves behind, with no edits of any kind. It passes $5$ of the $113$ scored
tasks. Four are core tasks, which it passes on every attempt:
\code{041\_platformer\_make\_checkpoints}, \code{052\_surburban\_trampoline\_bounce},
\code{059\_surburban\_merrygoround\_max\_velocity} and
\code{071\_surburban\_teleport\_sandbox}. The fifth is the debugging task
\code{010\_left\_shift\_sprint\_5s\_bug\_2}. The four core checks each assert a property of a
live play session that the untouched world already satisfies, such as a player coming to rest
inside a tolerance that the target already meets. We keep all four scored. Models still fail
three of them on $9$ to $22\%$ of attempts, so these three separate agents that break a world
that was already correct. The fourth, \code{059\_surburban\_merrygoround\_max\_velocity}, is
passed by all $13$ models on all $208$ attempts and separates nothing.

\section{Extended Related Work}
\label{app:related}

\textsc{HumanEval} and \textsc{MBPP} \citep{chen2021codex,austin2021mbpp} established
execution-based scoring against unit tests for code generation. \textsc{SWE-bench}
\citep{jimenez2024swebench} and its verified subset \citep{chowdhury2024verified} are
methodologically closest to our executable verification: they include real repository context,
require coordinating changes across files, and come with a reproducible harness. Our setting
differs in where the work happens and how it can be judged. A SWE-bench task is text in a file
tree, and the repository's own tests run over it once the agent stops. Our tasks allow two kinds
of action with different timing. Executing \lang{} code in the live engine session changes the
game scene immediately, while edited scripts show their behavior only once the game is running
and a player acts.

In \textsc{AppWorld} \citep{trivedi2024appworld}, agents write programs against a stateful
world of apps and must discover task-relevant facts through API calls. Its agents, however,
operate existing applications through a fixed API. Ours write the behavior that is later
exercised in the game.

\citet{mehtiyev2026beyond} ask whether an agent's investigative \emph{policy} adapts to task
difficulty, and find that it does not. We ask whether the \emph{association between that
behavior and the outcome} varies by kind of task, and we cannot detect a difference. The two
findings are consistent on a point that matters: agents apply a fixed investigative policy, even
though in both kinds of work we study, investigation pays off.

A growing literature documents contamination and weak construct validity in LLM
benchmarks \citep{bean2025constructvalidity,balloccu2024contamination}. The Agentic Benchmark
Checklist (ABC) \citep{zhu2025abc} formalizes task validity, outcome validity and benchmark
reporting, and asks that tasks be designed so that success cannot come from guessing.
\toolname{}'s checks are hand-written and human-verified against explicit reference solutions,
and together with its process metrics they are designed to meet these criteria.
\S\ref{sec:limitations} reports our known limitations.

A related line of work applies item-response theory to benchmarks, treating questions as items
with estimated difficulty and discrimination. \citet{kipnis2025metabench} distill six benchmarks
($28{,}632$ items) to under $3\%$ of their combined size, and \citet{maiapolo2024tiny} estimate full-benchmark scores from $100$-item subsets. Both fit item
parameters on far more respondents than we have: tens to hundreds of models for tinyBenchmarks and
thousands for metabench. We fit the same class of model in \S\ref{sec:frontier} to find which
of our tasks separate models. With only $13$ respondents, we report the tasks whose
discrimination is identified rather than compressing the suite.

Like \toolname{}, \textsc{Terminal-Bench} \citep{terminalbench2026} has agents write software
inside a stateful environment that they must explore and manipulate. In the authors' words,
however, its tests ``do not test the agent's commands or console output,'' a deliberate choice
for a purely outcome-driven benchmark. In \textsc{GameCraft-Bench} \citep{luo2026gamecraft},
agents build a launchable Godot game and submit replayable input traces, which a verifier
replays and a multimodal judge scores against a hidden rubric. The score depends only on the delivered game. The authors describe agent behavior, such as
tool-call counts and screenshot inspections, but the trajectory does not enter the score.

\section{Per-Model Results}
\label{app:leaderboard}

Table~\ref{tab:leaderboard} in \S\ref{sec:frontier} is the core-suite leaderboard behind the
main text's estimates. This appendix gives additional results, and Table~\ref{tab:debug} reports
the debugging suite. The $29$ debugging tasks form a separate suite and are never averaged into
the core results (\S\ref{sec:benchmark}).

\begin{figure}[tbp]
\centering
\includegraphics[width=0.70\linewidth]{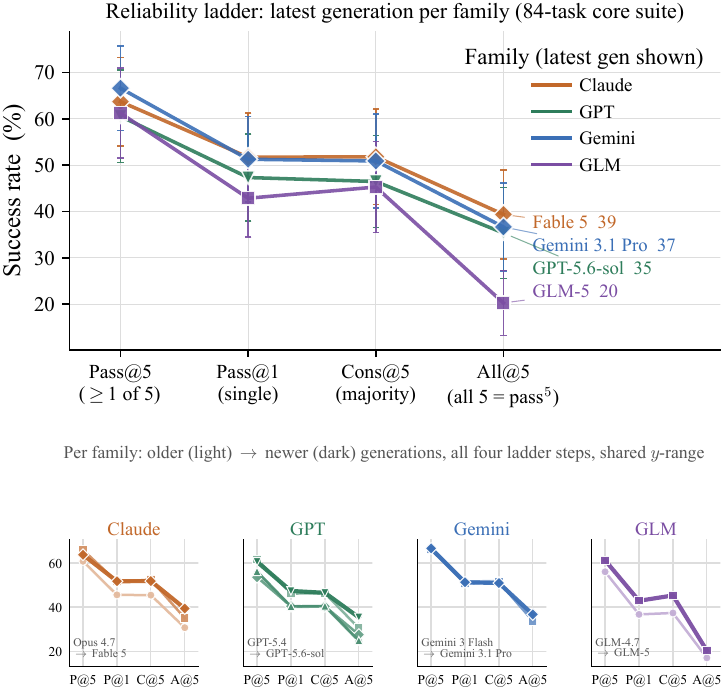}
\caption{The reliability ladder (84-task core suite, $k{=}16$). Each line traces one
model as the success criterion tightens from pass@5 ($\geq\!1$ of 5) through
pass@1 and cons@5 to all@5 (all 5, i.e.\ pass$^5$). The $\sim\!31$-point drop is
the consistency gap. Line crossings show that the reliability ranking is close to the single-attempt ranking
but not the same (Spearman $0.90$ between pass@1 and all@5 ranks in
Table~\ref{tab:leaderboard}). \emph{Main panel:} the latest generation of each
family. \emph{Small panels:} every generation within a family, showing that the
crossings persist within families.}
\label{fig:reliability}
\end{figure}

A growing reliability-science literature argues that accuracy alone does not describe how an
agent will behave in deployment, and that consistency across repeated attempts, which pass$^k$
(our all@5) measures, should be reported alongside it
\citep{khanal2026beyond,rabanser2026reliability}. Figure~\ref{fig:reliability} traces the
consistency gap: as the reliability bar tightens from pass@5 to all@5, success falls by
$\sim\!31$ points.

\begin{table}[tbp]
\centering
\caption{\textbf{The secondary suite: 29 debugging tasks}, scored on its own and never
averaged into Table~\ref{tab:leaderboard}. Columns are as there. \S\ref{sec:frontier} reports how widely the top five models spread on it.}
\label{tab:debug}
\footnotesize
\setlength{\tabcolsep}{4pt}
\begin{tabular}{@{}lccccccccc@{}}
\toprule
Model & Pass@1 & Pass@5 & Cons@5 & All@5 & Tools & Obs & Obs\% & Err (\%) & Tmo (\%) \\
\midrule
Claude Fable 5         & $\mathbf{65.5}_{(7.9)}$ & $80.4_{(8.2)}$ & $66.3_{(8.7)}$ & $\mathbf{48.8}_{(9.3)}$ & 5.8 & 4.1 & 71 & 2.53 & 2.37 \\
GPT-5.6-luna           & $65.1_{(7.6)}$ & $80.3_{(8.6)}$ & $\mathbf{68.2}_{(8.5)}$ & $43.2_{(8.5)}$ & 9.5 & 7.8 & 82 & 2.63 & 1.72 \\
Gemini 3.1 Pro         & $63.6_{(7.8)}$ & $\mathbf{80.5}_{(9.2)}$ & $64.9_{(8.8)}$ & $42.3_{(8.4)}$ & 7.7 & 4.8 & 63 & 4.65 & \textbf{0.00} \\
Claude Opus 4.8        & $62.7_{(8.2)}$ & $76.6_{(7.9)}$ & $64.2_{(9.0)}$ & $46.1_{(8.7)}$ & 7.1 & 4.8 & 68 & 7.42 & 1.94 \\
Claude Opus 4.7        & $59.1_{(7.9)}$ & $78.8_{(7.4)}$ & $57.9_{(9.0)}$ & $41.6_{(9.1)}$ & 5.6 & 4.0 & 72 & 1.95 & 2.16 \\
GLM-5                  & $59.1_{(7.8)}$ & $75.1_{(8.5)}$ & $63.0_{(9.3)}$ & $35.5_{(7.9)}$ & 6.5 & 5.0 & 76 & 1.89 & 1.72 \\
GPT-5.6-sol            & $58.8_{(8.5)}$ & $72.7_{(8.8)}$ & $59.2_{(9.1)}$ & $43.7_{(9.3)}$ & 11.4 & 9.7 & 86 & 8.79 & 3.45 \\
GPT-5.6-terra          & $58.8_{(7.7)}$ & $77.0_{(8.7)}$ & $61.1_{(8.4)}$ & $37.0_{(9.0)}$ & 8.3 & 6.5 & 79 & 3.33 & 2.37 \\
Gemini 3 Flash         & $57.3_{(8.3)}$ & $74.3_{(7.5)}$ & $56.8_{(9.4)}$ & $41.8_{(8.8)}$ & 12.6 & 9.3 & 74 & 3.75 & 2.16 \\
GPT-5.5                & $57.1_{(8.6)}$ & $68.1_{(9.7)}$ & $59.0_{(9.2)}$ & $42.5_{(9.3)}$ & 7.4 & 6.0 & 81 & \textbf{1.61} & 1.29 \\
Gemini 3.5 Flash       & $52.4_{(7.4)}$ & $73.8_{(8.5)}$ & $54.1_{(8.6)}$ & $26.0_{(6.7)}$ & 8.7 & 6.0 & 70 & 4.13 & 1.94 \\
GPT-5.4                & $51.3_{(7.8)}$ & $70.0_{(8.7)}$ & $51.0_{(8.6)}$ & $31.6_{(8.3)}$ & 9.6 & 8.2 & 85 & 2.04 & 2.16 \\
GLM-4.7                & $50.2_{(7.2)}$ & $74.7_{(7.8)}$ & $52.4_{(8.8)}$ & $21.3_{(6.2)}$ & 8.7 & 6.0 & 69 & 6.19 & 2.16 \\
\bottomrule
\end{tabular}
\end{table}

\paragraph{Timeout policy.}
We treat the time budget as part of the quality bar. Timeout rates are uneven. Gemini~3.1~Pro
and GLM-4.7 never time out on the core suite ($0\%$), GPT-5.4 times out on $3.6\%$ of attempts
($48$ of $1{,}344$), and four of the five leaders time out at least occasionally. The most
generous alternative is to drop timeouts instead of counting them as failures, which credits a
model with its success rate on the attempts that finished. This moves core pass@1 by at most
$1.5$pp for GPT-5.4 and by at most $1.2$pp for every other model. That is enough to change some
ranks but none of our claims. The top five remain the same five models and Claude~Fable~5 stays
first, though their order among themselves changes, as expected for aggregates that do not
separate (\S\ref{sec:frontier}). Further down, GPT-5.4 rises from twelfth to tenth, passing
GPT-5.6-terra and GPT-5.6-luna. Rank among nearly tied models therefore partly reflects
finishing within the budget, which is what we intend.

\section{Statistical Procedures}
\label{app:stats}

\paragraph{The panel.}
Every (model,\,task) cell carries exactly $k{=}16$ attempts, so every cell is estimated at the
same precision: $17{,}472$ core attempts over $1{,}092$ cells. No attempt is dropped from the
corpus on the basis of its outcome.

\paragraph{Resampling.}
Suite-level rates are reported with a studentized bootstrap over tasks, which captures both the
spread between tasks and the sampling variance within a task. The paired intervals behind every
model-versus-model claim use a separate estimator: a percentile bootstrap over tasks on the mean
per-task difference ($5{,}000$ resamples, seed $0$), without studentization. The Wilcoxon
signed-rank check of \S\ref{sec:metrics} agrees with this bootstrap on $75$ of the $78$ pairs.
The three pairs where they disagree all lie between $p{=}0.03$ and $p{=}0.08$, so no comparison
we report depends on the choice between them. For the paired estimator, the minimum detectable
effect at $84$ tasks, $\alpha{=}0.05$ two-sided and $80\%$ power, is $9.2$pp for the median model
pair and ranges from $6.2$ to $11.7$pp across the $78$ pairs. A marginal comparison, which
discards the pairing, needs $18.3$pp, which is why no pair separates marginally.

The count of resolved pairs in \S\ref{sec:frontier} uses $20{,}000$ resamples with a per-pair
seed. The Holm step-down at $\alpha{=}0.05$ across the $78$ pairs, reported in
\S\ref{sec:frontier}, is the only family-wise correction we make. We apply none within a single
paired comparison, because the quantities examined there are strongly dependent and each
conclusion rests on an interval rather than on clearing a threshold.

\paragraph{Clustering.}
Resampling tasks treats them as independent, but they are not fully independent. The $84$
core tasks are authored against $11$ shared place files, so tasks in the same place share a
scene graph and scripts. We therefore recomputed every suite interval with place files, not
tasks, as the resampling unit. The within-place intraclass correlation of per-task pass@1
averages $-0.000$ across the $13$ models, which implies a design effect of $1.00$, and the
clustered standard errors average $1.06$ times the task-level ones. No leaderboard conclusion
changes. The trajectory estimates of \S\ref{sec:grounded} cluster on tasks rather than
resampling them, and \S\ref{app:traj} recomputes each of them in three further ways.

\section{Item-Response Model Specification}
\label{app:irt}

This appendix gives the full specification of the 2PL fit reported in \S\ref{sec:frontier}, its
diagnostics, and a prior sensitivity analysis. In short, the model's \emph{scale} depends on the
prior but its \emph{orderings} do not, which is why the main text relies on rankings of $\alpha$
and $\beta$ rather than their magnitudes.

\paragraph{Data and likelihood.}
The response matrix is the $13$ models $\times\,84$ core tasks of the core panel. Cell
$(m,e)$ contributes $n_{me}{=}16$ attempts and $c_{me}$ passes. Writing $\theta_m$ for
model ability, $\beta_e$ for task difficulty and $\alpha_e$ for task discrimination,
\[
y_{me}\;\sim\;\mathrm{Binomial}\!\left(n_{me},\,p_{me}\right),
\qquad
p_{me}\;=\;\operatorname{logit}^{-1}\!\big(\alpha_e\,(\theta_m-\beta_e)\big).
\]

\paragraph{Hierarchical structure and priors.}
Task parameters are partially pooled, and so are model abilities, which are nested in the four
provider families $f(m)\in\{$Claude, Gemini, GPT, GLM$\}$. All three levels use a non-centered parameterization, which makes the funnel geometry
sampleable:
\[
\begin{aligned}
\theta_m &= \sqrt{\rho}\,\tilde\phi_{f(m)} + \sqrt{1-\rho}\,\tilde\theta_m,
  &\tilde\phi_f,\ \tilde\theta_m,\ \tilde\beta_e,\ \tilde\alpha_e &\sim \mathcal{N}(0,1),
  &\rho &\sim \mathrm{Beta}(1,1), \\
\beta_e &= \mu_\beta + \sigma_\beta\,\tilde\beta_e,
  &\mu_\beta &\sim \mathcal{N}(0,1.5),
  &\sigma_\beta &\sim \mathrm{HalfNormal}(0.5), \\
\log\alpha_e &= \mu_{\log\alpha} + \sigma_{\log\alpha}\,\tilde\alpha_e,
  &\mu_{\log\alpha} &\sim \mathcal{N}(0,0.5),
  &\sigma_{\log\alpha} &\sim \mathrm{HalfNormal}(1.0).
\end{aligned}
\]
The $13$ models are not independent respondents. They come from four families (three Claude,
three Gemini, five GPT and two GLM), and versions within a family share training data, tooling
and post-training. Treating them as $13$ exchangeable students would understate how correlated
their responses are, and the family level above handles this.

Estimating $\alpha_e$ on the log scale forces $\alpha_e{>}0$, so every item characteristic
curve increases with ability, and by construction no task can be easier for a weaker model. This
is a modeling assumption, not a finding.

\paragraph{Identification.}
The likelihood depends on the parameters only through $\alpha_e(\theta_m-\beta_e)$, so the
location and scale of the latent axis and the scale of $\alpha$ are jointly unidentified. The
usual 2PL convention anchors them by fixing $\theta_m\sim\mathcal{N}(0,1)$ with no estimated
hyperparameters. The parameterization above keeps that anchor exactly: $\tilde\phi$ and
$\tilde\theta$ are standard normal and the weights are $\sqrt\rho$ and $\sqrt{1-\rho}$, so
$\theta_m$ has total variance $1$ whatever $\rho$ is. Adding the family level therefore costs nothing in identification, and it lets us estimate the
share of ability variance that lies between families rather than within them.

We compute that share from the fitted abilities. On each posterior draw, we split each model's
ability into its family mean and its deviation from that mean, and take the variance of the
family means as a fraction of the total. This gives $0.792$ $[0.707,0.871]$: about four fifths
of the spread in ability lies between providers rather than between versions within a provider.
Two models from the same family resemble each other far more than two models from different
families, which is why we do not treat the $13$ models as independent respondents.

The family level does not create this concentration. Refitting with no family
structure at all, $\theta_m\sim\mathcal{N}(0,1)$ independently across models, still leaves
$0.762$ $[0.670,0.844]$ of the ability variance between families. Permuting the family labels
at random drops it to $0.247$ $[0.026,0.593]$ ($p{=}0.002$).

\paragraph{Continuity correction.}
Observed successes enter as $\min(n-\tfrac12,\max(\tfrac12,c))$. Without this correction, the $54.7\%$ of
cells sitting exactly at $0$ or $n$ would push the corresponding $\beta_e$ toward $\pm\infty$.
This is a deliberate bias toward the interior: it shrinks the most extreme difficulties, and
it is one reason we do not interpret $\beta$ magnitudes.

\paragraph{Sampling and diagnostics.}
NUTS via \texttt{numpyro}, $4$ chains, $2{,}000$ tuning iterations and $1{,}000$ retained
draws each ($4{,}000$ posterior draws), \texttt{target\_accept}${=}0.9$, seed $42$. The fit
reports $5$ divergent transitions, $\hat R\le1.000$ across $\theta$, $\beta$ and $\alpha$
($1.010$ over all parameters including the raw non-centered terms), and minimum bulk/tail
effective sample size $1{,}237$. Posterior means (sd) for the hyperparameters are
$\mu_\beta{=}0.25\,(0.49)$, $\sigma_\beta{=}1.50\,(0.31)$,
$\mu_{\log\alpha}{=}0.65\,(0.24)$, $\sigma_{\log\alpha}{=}0.86\,(0.10)$.

\paragraph{Prior sensitivity.}
$\sigma_\beta$ has a posterior mean of $1.50$ against a prior median of $0.34$. The data call
for a much wider spread of task difficulties than $\mathrm{HalfNormal}(0.5)$ expects, and the
prior is actively shrinking it. Because only the product $\alpha_e(\theta_m-\beta_e)$ is
identified, compressing the difficulty scale inflates the discrimination scale to compensate.
The panel shows why the data pull so hard: $54.7\%$ of the $1{,}092$ model-by-task cells are
all-pass or all-fail. Such a cell pushes difficulty toward infinity in one direction, and only
the prior keeps $\beta$ finite.

We therefore refit with diffuse hyperpriors on both item parameters, loosening both scales and
both locations rather than only the difficulty scale: $\sigma_\beta\sim\mathrm{HalfNormal}(10)$,
$\sigma_{\log\alpha}\sim\mathrm{HalfNormal}(3)$, $\mu_\beta\sim\mathcal{N}(0,10^2)$ and
$\mu_{\log\alpha}\sim\mathcal{N}(0,3^2)$. Ability keeps total variance $1$ in both fits, because
that fixes the scale of an otherwise unidentified model. The diffuse fit still converges ($16$
divergences, $\hat{R}\le1.010$, minimum ESS $1{,}161$). The two item parameters trade off exactly
as expected: $\sigma_\beta$ rises to $8.04$, the range of $\beta$ widens from $[-2.2,+3.1]$ to
$[-11.1,+16.0]$ logits, the median $\alpha$ falls from $1.96$ to $0.42$, and the maximum
$\alpha$ falls from $13.0$ to $2.8$.

None of our IRT-based claims changes substantially under the refit. The rank order of tasks is
preserved on both parameters (Spearman $0.999$ for each). The refit selects the same seven tasks
that \S\ref{sec:frontier} reports, because the criterion is relative to each fit's own median
$\alpha$ rather than an absolute threshold. An absolute cut would select all seven under the
reported priors and none under the diffuse ones, only because the scale of $\alpha$ shrinks by a
factor of about five. The between-family share of ability is also unchanged: $0.793$
$[0.712,0.880]$ against $0.792$ $[0.707,0.871]$.

\paragraph{Six unsolved tasks and $\alpha$.}
The six core tasks that no model solves are \code{019\_secret\_door\_puzzle},
\code{021\_gem\_orbiting\_part}, \code{035\_laser\_tag\_regenerate\_health},
\code{057\_surburban\_no\_trespassing}, \code{073\_lasertag\_crate\_drop\_disappear} and
\code{115\_racing\_wrong\_direction\_warning}. The four that every model solves on every attempt
are \code{013\_inf\_cube\_fall}, \code{059\_surburban\_merrygoround\_max\_velocity},
\code{083\_platformer\_coin\_increment\_down} and \code{113\_racing\_extend\_laps}.
For a task that no model solves, the data only establish that its response probability is near
zero for every model, and any sufficiently large $\beta_e$ achieves that at any $\alpha_e$. The
two parameters therefore trade off along a ridge instead of being separately identified, and
where the posterior lands on that ridge depends on how far the $\sigma_\beta$ prior lets $\beta$
travel. Their parameters are therefore not interpretable. Under the reported priors, the six
tasks sit at $\beta{\approx}2.3$ with $\alpha{\approx}6.3$. Under the diffuse prior they move to
$\beta{\approx}11.8$ with $\alpha{\approx}1.3$. Under both settings, their $\alpha$ credible
intervals are about $4.6$ times wider than those of other tasks, and all six rank between $6$th
and $12$th of $84$ by $\alpha$. Whichever prior is used, six of the twelve largest $\alpha$
values belong to tasks that carry no between-model information. This distortion of the $\alpha$
ranking is a property of the design, not of the prior.

\section{Item Analysis Details}
\label{app:items}

Table~\ref{tab:disc} lists the seven tasks whose discrimination \S\ref{sec:frontier} reports as
clearly above the suite median.

\begin{table}[h]
\centering
\caption{\textbf{Where discrimination is identified.} The seven core tasks whose $95\%$
interval for discrimination $\alpha$ lies above the suite's median $\alpha$ of
$1.96$. Because $\alpha$'s level is not identified independently of ability, the criterion is
relative to each fit's own $\alpha$ scale. We select these same seven under the
diffuse-hyperprior refit of \S\ref{app:irt}, whose $\alpha$ values are roughly five times
smaller. Difficulty $\beta$ is on the ability logit scale, across which the $84$ tasks span
$\beta\in[-2.2,+3.1]$.}
\label{tab:disc}
\small
\setlength{\tabcolsep}{5pt}
\begin{tabular}{@{}llrlrr@{}}
\toprule
Task & Kind & $\alpha$ & $95\%$ interval & $\beta$ & pass@1 \\
\midrule
\code{103\_city\_lights\_on\_off}           & script & $12.97$ & $[5.27,22.91]$ & $+0.27$ & $14.4\%$ \\
\code{092\_fps\_shoot\_ground\_bounce}      & script & $10.67$ & $[4.23,17.50]$ & $+0.24$ & $19.2\%$ \\
\code{020\_gravity\_well}                   & script & $7.10$  & $[3.04,11.92]$ & $-0.30$ & $82.2\%$ \\
\code{105\_lasertag\_friendly\_fire}        & script & $6.75$  & $[3.22,11.28]$ & $-0.27$ & $79.3\%$ \\
\code{049\_surburban\_fridge\_door\_open}   & script & $6.05$  & $[3.00,9.70]$  & $-0.15$ & $66.8\%$ \\
\code{108\_racing\_fix\_getdriversinpart}   & script & $5.47$  & $[2.13,9.21]$  & $-0.46$ & $89.9\%$ \\
\code{054\_surburban\_equip\_flashlight}    & scene  & $4.78$  & $[1.97,8.10]$  & $-0.39$ & $83.2\%$ \\
\bottomrule
\end{tabular}
\end{table}

\section{Trajectory Analysis Robustness}
\label{app:traj}

This appendix states which tasks each row of Table~\ref{tab:grounded} is estimated on, defines
the features the main text does not, and reports alternatives to the inference scheme and the
linear model. Of the $84$ core tasks, $54$ carry the instance annotation that coverage is built
from, and coverage is defined on exactly those $54$. The other $30$ are dropped from the coverage
rows, although $7$ of them carry the script-dependency annotation and feed that channel instead.
Of the $54$, the $37$ whose required tools place them clearly in one kind of work form the two
arms of \S\ref{sec:grounded}: $24$ scene-only and $13$ script-only. The other $17$ require both
tools and are reported separately below. No annotation is shown to the model, and none is used
in grading. Cluster counts for every row are listed under the inference alternatives below.

\paragraph{Features in Table~\ref{tab:grounded}.}
The two sequencing features in the table's lower block are (i) whether the run followed a
scene-graph or text search by reading a script and (ii) the fraction of script reads whose
target had already appeared in an earlier observation result rather than being guessed. Neither
uses any annotation. The reference-script rows appear only for the script arm, because just $2$
scene tasks carry that annotation, too few to estimate a difference.

\paragraph{Tasks that require both tools.}
The partition excludes these tasks on purpose. Splitting by a requirement and its complement
would put the $29$ core tasks that need both \code{execute\_luau} \emph{and} \code{multi\_edit}
into the script arm, making it half scene work. We therefore score only the two classes that
share no required tool. The excluded group is uninformative rather than contradictory: on its
$17$ annotated tasks, coverage gives $-2.8$pp none$\to$full, with an interval too wide to
interpret ($[-21.6,+15.9]$).

\paragraph{Non-circularity.}
On the scene arm, per standard deviation, the estimate depends on how strictly a dependency is
credited. Crediting a dependency reached through an action as well as an observation gives
$+14.2$pp. Crediting only observation, at any point in the run, gives $+7.8$pp. Crediting only
observation that precedes the run's first action on that dependency, which is what we report,
gives $+6.4$pp. A still stricter variant requires the observation to precede the run's first
action of any kind. It gives $+3.9$pp $[-0.1,+7.9]$ on the scene arm and $+3.0$pp $[+0.3,+5.8]$
on the script arm. We do not headline it, because \code{execute\_luau} is both the graded action
on scene tasks and the only way to run engine code that probes the live world. Cutting the window
at the first such call discards observation that is legitimately interleaved with probing.

\paragraph{Additional controls.}
We also run several robustness checks. The first two are on the none$\to$full scale and compare
against the $+13.4$pp of \S\ref{sec:grounded}. Controlling for the number of distinct scripts
read leaves the coefficient at $+13.4$pp. The place file a task is set in does not change the
result, though this check is weaker than the others. Two of the $24$ scene tasks are authored
against a bare baseplate, and dropping them leaves the estimate essentially unchanged at
$+13.0$pp, but with only $22$ clusters left the interval no longer excludes zero
($[-0.9,+26.9]$). Dropping the tasks the benchmark labels as hard leaves the scene estimate
unchanged at $+6.4$pp per standard deviation and the script estimate at $+3.3$pp
$[+0.4,+6.1]$. This restriction affects only the script arm, because no scene-change task
carries the hard label. Finally, counting timeouts as failures, as the leaderboard does, rather
than dropping them moves every estimate by at most $0.1$pp, so the signal does not depend on
conditioning on a graded verdict.

\paragraph{Alternative inference schemes.}
Each row of Table~\ref{tab:grounded} uses every task in the relevant class on which its feature
is defined, so the sample differs from row to row. The coverage rows are limited to tasks with a
dependency annotation and the expected-tool row to tasks with an expected-tool annotation. The
two sequencing rows need no annotation, so they run on the full classes of $25$ scene-only and
$30$ script-only tasks. The intervals rest on few clusters (scene/script): $24/13$ tasks for
observing reference instances, $11$ for reading reference scripts, $13/26$ for expected-tool
match, $25/30$ for search-then-read, and $20/30$ for grounded reads. With so few clusters, the
cluster-robust sandwich of \S\ref{sec:grounded}, which applies a $G/(G{-}1)$ correction for $G$
clusters and uses a normal reference distribution, can under-cover. We therefore recomputed
every row in three further ways: the same sandwich against a $t(G{-}1)$ reference, a score wild
cluster bootstrap by task with Rademacher weights ($999$ draws), and two-way clustering on task
and model \citep{cameron2011robust}. No entry changes its verdict under any of them. The
headline row's interval is $[+0.6,+12.3]$ as reported, $[+0.3,+12.6]$ under $t(23)$,
$[+1.1,+11.7]$ under the bootstrap, and $[+0.5,+12.3]$ under two-way clustering. 

\begin{figure}[tbp]
\centering
\includegraphics[width=0.80\linewidth]{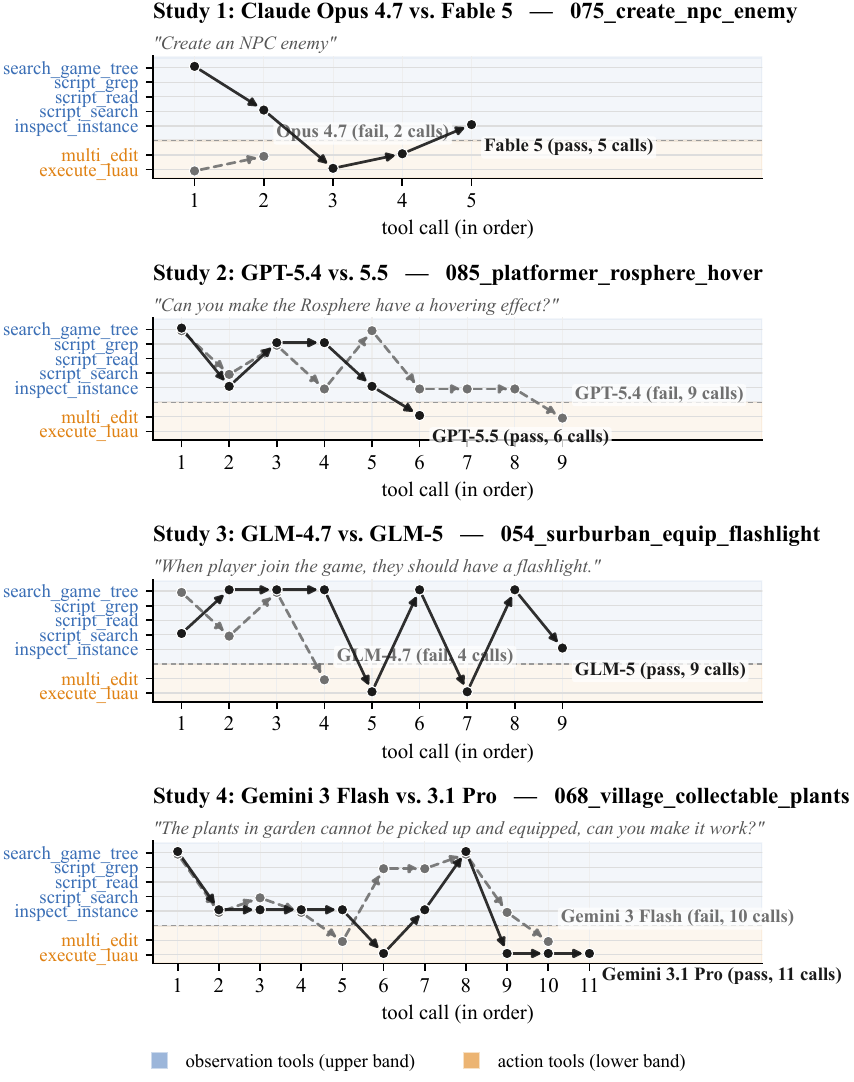}
\caption{Representative per-run tool-call trajectories. The vertical axis lists seven of the eight tools (all but \code{insert\_from\_creator\_store}), grouped into an observation band (upper, blue) and an action band
(lower, orange). The older model's path is grey and dashed, and the newer model's is black and solid. Arrows
trace the order of calls, so a path that stays high keeps gathering information and a path that
drops low is acting. Each panel compares runs of an older and a newer model on a task whose
\emph{outcome} differs between them, with the newer model passing in every case:
(1)~\code{075\_create\_npc\_enemy}, Opus~4.7 to Fable~5; (2)~\code{085\_platformer\_rosphere\_hover},
GPT-5.4 to 5.5; (3)~\code{054\_surburban\_equip\_flashlight}, GLM-4.7 to GLM-5;
(4)~\code{068\_village\_collectable\_plants}, Gemini~3~Flash to 3.1~Pro. Not every flip looks
like this. Some pairs trace nearly identical short paths and differ only in outcome, so the
trajectory adds a signal but does not fully explain the flip.}

\label{fig:traj}
\end{figure}

\clearpage
\section{The Scored Tasks}
\label{app:catalog}

Every scored task is listed below with the prompt the agent receives, the place file the
session opens, and the check phases the task uses. \code{check\_scene} inspects the edited world
at edit time, and \code{check\_game} asserts a property of a play session started after the
edits. A task passes only if every phase it defines passes. A task with only one phase marked is
not weaker: its objective is observable in only one of the two places. Prompts are reproduced
verbatim, with whitespace collapsed. One prompt is cut at $400$ characters (marked
\code{[...]}), and a serialization wrapper is stripped from one other.

\begin{center}
\begin{footnotesize}
\begin{longtable}{@{}%
>{\raggedright\arraybackslash}p{0.190\linewidth}%
>{\raggedright\arraybackslash}p{0.120\linewidth}%
@{\hspace{5pt}}c@{\hspace{5pt}}c@{\hspace{7pt}}%
>{\raggedright\arraybackslash}p{0.440\linewidth}@{}}
\caption{The 84 core tasks. Scene and Play mark which check phases the task defines: \code{check\_scene} at edit time and \code{check\_game} in a play session.}\label{tab:catalog-core}\\
\toprule
Task & Place & Scene & Play & Prompt \\
\midrule
\endfirsthead
\multicolumn{5}{@{}l}{\emph{Table~\ref{tab:catalog-core}, continued.}}\\
\toprule
Task & Place & Scene & Play & Prompt \\
\midrule
\endhead
\bottomrule
\endfoot
\texttt{001\_\allowbreak{}make\_\allowbreak{}cars\_\allowbreak{}faster} & \texttt{racing} & $\bullet$ & \,--\, & Make the cars of this game 2x faster \\
\texttt{002\_\allowbreak{}emit\_\allowbreak{}white\_\allowbreak{}smoke} & \texttt{surburban} & $\bullet$ & \,--\, & Make white smoke come out of every chimney. \\
\texttt{003\_\allowbreak{}make\_\allowbreak{}leaves\_\allowbreak{}fall\_\allowbreak{}colored} & \texttt{village} & $\bullet$ & \,--\, & Make the leaves of trees in this game fall colored \\
\texttt{004\_\allowbreak{}reduce\_\allowbreak{}car\_\allowbreak{}friction\_\allowbreak{}enable\_\allowbreak{}sliding} & \texttt{racing} & $\bullet$ & \,--\, & Reduce the friction of the cars that spawn in this game with the ground so I can slide around \\
\texttt{007\_\allowbreak{}change\_\allowbreak{}logo\_\allowbreak{}cube\_\allowbreak{}green} & \texttt{platformer} & $\bullet$ & \,--\, & Change the Roblox logo like cube to be green \\
\texttt{008\_\allowbreak{}spawn\_\allowbreak{}as\_\allowbreak{}r6} & \texttt{baseplate} & $\bullet$ & $\bullet$ & make every player spawn as a R6 test dummy as their character for now. \\
\texttt{010\_\allowbreak{}left\_\allowbreak{}shift\_\allowbreak{}sprint\_\allowbreak{}5s} & \texttt{baseplate} & \,--\, & $\bullet$ & Can you make a script where if you hold the left shift button your walkspeed gets 20 added to it and if you hold it for 5 seconds your walkspeed goes back to normal? \\
\texttt{011\_\allowbreak{}change\_\allowbreak{}height\_\allowbreak{}to\_\allowbreak{}1} & \texttt{baseplate} & \,--\, & $\bullet$ & chahge players height to 1 \\
\texttt{012\_\allowbreak{}remove\_\allowbreak{}legs} & \texttt{baseplate} & \,--\, & $\bullet$ & Make a script so that the game removes your 2 legs \\
\texttt{013\_\allowbreak{}inf\_\allowbreak{}cube\_\allowbreak{}fall} & \texttt{baseplate} & \,--\, & $\bullet$ & make it so that a size 1,1,1 cube falls at position 0,0,0 every second \\
\texttt{017\_\allowbreak{}make\_\allowbreak{}traffic\_\allowbreak{}light} & \texttt{baseplate} & \,--\, & $\bullet$ & write me a script for a simple 4 way traffic light \\
\texttt{018\_\allowbreak{}weather\_\allowbreak{}machine} & \texttt{baseplate} & $\bullet$ & $\bullet$ & Create a 'WeatherMachine' part that, when activated, cycles through sunny, rainy, and foggy conditions every 3 seconds, visually affecting the entire game world. \\
\texttt{019\_\allowbreak{}secret\_\allowbreak{}door\_\allowbreak{}puzzle} & \texttt{baseplate} & \,--\, & $\bullet$ & Design a puzzle, when stepping on color tile parts in the order of red, blue, green, yellow, the secret door will open. \\
\texttt{020\_\allowbreak{}gravity\_\allowbreak{}well} & \texttt{baseplate} & \,--\, & $\bullet$ & Add a 'GravityWell' part that, when touched, inverts a player's gravity for 5 seconds or until they leave its spherical influence. \\
\texttt{021\_\allowbreak{}gem\_\allowbreak{}orbiting\_\allowbreak{}part} & \texttt{baseplate} & $\bullet$ & $\bullet$ & Make a gem orbit around this part, and automatically attract towards the player when they are within 10 studs. \\
\texttt{022\_\allowbreak{}add\_\allowbreak{}world\_\allowbreak{}time} & \texttt{baseplate} & \,--\, & $\bullet$ & Add a script that adds 10 seconds to the world time in Lighting every second. \\
\texttt{023\_\allowbreak{}music\_\allowbreak{}playing\_\allowbreak{}part} & \texttt{baseplate} & \,--\, & $\bullet$ & write me a script to play and stop music when within certain trigger parts. \\
\texttt{025\_\allowbreak{}chase\_\allowbreak{}and\_\allowbreak{}damage} & \texttt{baseplate} & \,--\, & $\bullet$ & make an entity that chase and damage the player \\
\texttt{026\_\allowbreak{}make\_\allowbreak{}traffic\_\allowbreak{}light\_\allowbreak{}v2} & \texttt{surburban} & \,--\, & $\bullet$ & write me a script for a simple 4 way traffic light \\
\texttt{026\_\allowbreak{}make\_\allowbreak{}traffic\_\allowbreak{}light\_\allowbreak{}v3} & \texttt{surburban} & \,--\, & $\bullet$ & write me a script for a simple 4 way traffic light \\
\texttt{027\_\allowbreak{}firstperson\_\allowbreak{}block} & \texttt{baseplate} & \,--\, & $\bullet$ & hey can you make me a block that makes you firstperson when you touch it? \\
\texttt{029\_\allowbreak{}play\_\allowbreak{}idle\_\allowbreak{}animation} & \texttt{baseplate} & \,--\, & $\bullet$ & generate a script that makes you play this animation everytime your idle in 5 seconds after your still idle: 507771019 \\
\texttt{030\_\allowbreak{}play\_\allowbreak{}idle\_\allowbreak{}animation\_\allowbreak{}v2} & \texttt{platformer} & \,--\, & $\bullet$ & generate a script that makes you play this animation everytime your idle in 5 seconds after your still idle: 507771019 \\
\texttt{033\_\allowbreak{}count\_\allowbreak{}coins} & \texttt{platformer} & $\bullet$ & \,--\, & count the coins in the game, and create same number of parts \\
\texttt{035\_\allowbreak{}laser\_\allowbreak{}tag\_\allowbreak{}regenerate\_\allowbreak{}health} & \texttt{laser\_\allowbreak{}tag} & \,--\, & $\bullet$ & Implement a health regeneration system that starts 2 seconds after taking damage and regenerates at 10 health per second. \\
\texttt{038\_\allowbreak{}platformer\_\allowbreak{}coin\_\allowbreak{}multiple\_\allowbreak{}pickup} & \texttt{platformer} & $\bullet$ & $\bullet$ & Make it so that the coins can be picked up multiple times. \\
\texttt{041\_\allowbreak{}platformer\_\allowbreak{}make\_\allowbreak{}checkpoints} & \texttt{platformer} & \,--\, & $\bullet$ & Make this a checkpoint when player step on it. \\
\texttt{043\_\allowbreak{}platformer\_\allowbreak{}bouncing\_\allowbreak{}jumper} & \texttt{platformer} & \,--\, & $\bullet$ & Make all OneJump become BounceJump, players will bounce when step on it. \\
\texttt{048\_\allowbreak{}surburban\_\allowbreak{}fountain\_\allowbreak{}insert} & \texttt{surburban} & $\bullet$ & \,--\, & Place the water fountain on backyard deck and fit the size there. \\
\texttt{049\_\allowbreak{}surburban\_\allowbreak{}fridge\_\allowbreak{}door\_\allowbreak{}open} & \texttt{surburban} & \,--\, & $\bullet$ & Change the store fridge door so that it can be opened with touch, not click. \\
\texttt{050\_\allowbreak{}surburban\_\allowbreak{}gaspump\_\allowbreak{}explode} & \texttt{surburban} & \,--\, & $\bullet$ & Gas pump explode when player is in NearArea. \\
\texttt{052\_\allowbreak{}surburban\_\allowbreak{}trampoline\_\allowbreak{}bounce} & \texttt{surburban} & \,--\, & $\bullet$ & Make people keep bouncing when they are on the trampoline. \\
\texttt{053\_\allowbreak{}surburban\_\allowbreak{}billboard\_\allowbreak{}change\_\allowbreak{}decal} & \texttt{surburban} & \,--\, & $\bullet$ & Make the billboard change image every 1 second, using the decals starting with 'ad'. \\
\texttt{054\_\allowbreak{}surburban\_\allowbreak{}equip\_\allowbreak{}flashlight} & \texttt{surburban} & \,--\, & $\bullet$ & When player join the game, they should have a flashlight. \\
\texttt{055\_\allowbreak{}surburban\_\allowbreak{}tree\_\allowbreak{}fallcolor\_\allowbreak{}approach} & \texttt{surburban} & \,--\, & $\bullet$ & Make trees change color to red when player approach it. \\
\texttt{057\_\allowbreak{}surburban\_\allowbreak{}no\_\allowbreak{}trespassing} & \texttt{surburban} & \,--\, & $\bullet$ & No trespassing: when player enters the fenced yard, they will receive 10 damage. \\
\texttt{058\_\allowbreak{}surburban\_\allowbreak{}merrygoround\_\allowbreak{}no\_\allowbreak{}fling} & \texttt{surburban} & \,--\, & $\bullet$ & Can you make the MerryGoRound not fling the player? \\
\texttt{059\_\allowbreak{}surburban\_\allowbreak{}merrygoround\_\allowbreak{}max\_\allowbreak{}velocity} & \texttt{surburban} & \,--\, & $\bullet$ & User's speed on MerryGoRound should cap at 1. \\
\texttt{068\_\allowbreak{}village\_\allowbreak{}collectable\_\allowbreak{}plants} & \texttt{village} & \,--\, & $\bullet$ & The plants in garden cannot be picked up and equipped, can you make it work? \\
\texttt{070\_\allowbreak{}village\_\allowbreak{}make\_\allowbreak{}npc\_\allowbreak{}walk} & \texttt{village} & \,--\, & $\bullet$ & Make NPCs walk around randomly \\
\texttt{071\_\allowbreak{}rainbow\_\allowbreak{}hexagon} & \texttt{platformer} & \,--\, & $\bullet$ & Make 6 parts in a hexagon in the colors of a rainbox, only the red part should kill the player when they touch it \\
\texttt{071\_\allowbreak{}surburban\_\allowbreak{}teleport\_\allowbreak{}sandbox} & \texttt{surburban} & \,--\, & $\bullet$ & When player step on the sandbox, they will be teleported to the spawn location. \\
\texttt{073\_\allowbreak{}lasertag\_\allowbreak{}crate\_\allowbreak{}drop\_\allowbreak{}disappear} & \texttt{laser\_\allowbreak{}tag} & \,--\, & $\bullet$ & Make each crate drop an item and disappear after shooting them. \\
\texttt{074\_\allowbreak{}red\_\allowbreak{}grass\_\allowbreak{}sway} & \texttt{flat\_\allowbreak{}terrain} & $\bullet$ & \,--\, & Make the grass red and sway from side to side \\
\texttt{074\_\allowbreak{}village\_\allowbreak{}fire\_\allowbreak{}pit\_\allowbreak{}damage} & \texttt{village} & \,--\, & $\bullet$ & Make all fire pits in the game damage the player when walked over \\
\texttt{075\_\allowbreak{}create\_\allowbreak{}npc\_\allowbreak{}enemy} & \texttt{laser\_\allowbreak{}tag} & $\bullet$ & \,--\, & Create an NPC enemy \\
\texttt{075\_\allowbreak{}village\_\allowbreak{}remove\_\allowbreak{}tutorial\_\allowbreak{}assets} & \texttt{village} & $\bullet$ & \,--\, & Remove tutorial assets from the map \\
\texttt{076\_\allowbreak{}plane\_\allowbreak{}flyby} & \texttt{baseplate} & \,--\, & $\bullet$ & Add a plane flyby scene \\
\texttt{079\_\allowbreak{}platformer\_\allowbreak{}roblonk\_\allowbreak{}blue\_\allowbreak{}raise} & \texttt{platformer} & \,--\, & $\bullet$ & Make the Roblonk Model blue and raise by 5 studs smoothly when touchiung it, and then back to normal when you stop touching it \\
\texttt{080\_\allowbreak{}surburban\_\allowbreak{}school\_\allowbreak{}lights\_\allowbreak{}on} & \texttt{surburban} & $\bullet$ & \,--\, & Make every light in the school start in the 'on' state. \\
\texttt{081\_\allowbreak{}racing\_\allowbreak{}car\_\allowbreak{}color\_\allowbreak{}change} & \texttt{racing} & $\bullet$ & $\bullet$ & Please make the cars of this game constantly change colors when they have an occupied driver. \\
\texttt{082\_\allowbreak{}platformer\_\allowbreak{}moving\_\allowbreak{}platform\_\allowbreak{}speed\_\allowbreak{}up} & \texttt{platformer} & $\bullet$ & \,--\, & Make the moving platforms speed up by 5. \\
\texttt{083\_\allowbreak{}platformer\_\allowbreak{}coin\_\allowbreak{}increment\_\allowbreak{}down} & \texttt{platformer} & \,--\, & $\bullet$ & Make the coins go down by 1 instead of up by 1 when collecting them. Don't let it go below 0. \\
\texttt{084\_\allowbreak{}platformer\_\allowbreak{}roblonk\_\allowbreak{}rotate} & \texttt{platformer} & $\bullet$ & \,--\, & Can you rotate roblonk by 90 degrees? \\
\texttt{085\_\allowbreak{}platformer\_\allowbreak{}rosphere\_\allowbreak{}hover} & \texttt{platformer} & \,--\, & $\bullet$ & Can you make the Rosphere have a hovering effect? \\
\texttt{086\_\allowbreak{}racing\_\allowbreak{}car\_\allowbreak{}jump} & \texttt{racing} & \,--\, & $\bullet$ & Make it so that when a player is driving a car and jumps, the car jumps \\
\texttt{088\_\allowbreak{}surburban\_\allowbreak{}garage\_\allowbreak{}door\_\allowbreak{}speed\_\allowbreak{}up} & \texttt{surburban} & $\bullet$ & \,--\, & Double the speed of the garage doors opening. \\
\texttt{089\_\allowbreak{}fps\_\allowbreak{}box\_\allowbreak{}fling\_\allowbreak{}harder} & \texttt{fps\_\allowbreak{}system} & $\bullet$ & \,--\, & Make the boxes fling harder when you shoot them. \\
\texttt{090\_\allowbreak{}fps\_\allowbreak{}display\_\allowbreak{}target\_\allowbreak{}damage\_\allowbreak{}ui} & \texttt{fps\_\allowbreak{}system} & \,--\, & $\bullet$ & Make the targets display the damage with a UI after shooting the targets. \\
\texttt{091\_\allowbreak{}surburban\_\allowbreak{}fix\_\allowbreak{}grass\_\allowbreak{}in\_\allowbreak{}house} & \texttt{surburban} & $\bullet$ & \,--\, & Fix the grass in the floor of the house named 'Double Story Urban House' which shouldnt be there \\
\texttt{092\_\allowbreak{}fps\_\allowbreak{}shoot\_\allowbreak{}ground\_\allowbreak{}bounce} & \texttt{fps\_\allowbreak{}system} & \,--\, & $\bullet$ & Make the player bounce in the air when shooting the ground within 5 studs of the player. \\
\texttt{093\_\allowbreak{}surburban\_\allowbreak{}jeep\_\allowbreak{}in\_\allowbreak{}every\_\allowbreak{}garage} & \texttt{surburban} & $\bullet$ & \,--\, & Place a copy of the jeep in every garage. \\
\texttt{094\_\allowbreak{}village\_\allowbreak{}add\_\allowbreak{}cave} & \texttt{village} & \,--\, & $\bullet$ & Add an enclosed cave to the map. The cave should have a ceiling, floor, and walls forming a room-like structure that a player can walk into. Don't do it on terrain. \\
\texttt{095\_\allowbreak{}racing\_\allowbreak{}nitro\_\allowbreak{}effect\_\allowbreak{}color} & \texttt{racing} & \,--\, & $\bullet$ & Make the nitro effect color the same color as the car. \\
\texttt{096\_\allowbreak{}fps\_\allowbreak{}target\_\allowbreak{}overhead\_\allowbreak{}health\_\allowbreak{}ui} & \texttt{fps\_\allowbreak{}system} & $\bullet$ & $\bullet$ & Make the targets have an overhead UI displaying their current health. \\
\texttt{097\_\allowbreak{}ugc\_\allowbreak{}mannequin\_\allowbreak{}match\_\allowbreak{}outfit} & \texttt{ugc\_\allowbreak{}homestore} & \,--\, & $\bullet$ & Create a new mannequin for every player currently in the game and make it match their outfit \\
\texttt{099\_\allowbreak{}city\_\allowbreak{}add\_\allowbreak{}cars} & \texttt{modern\_\allowbreak{}city} & $\bullet$ & \,--\, & Add cars on the road \\
\texttt{100\_\allowbreak{}obby\_\allowbreak{}add\_\allowbreak{}death\_\allowbreak{}trap} & \texttt{classic\_\allowbreak{}obby} & $\bullet$ & $\bullet$ & Add a spinning death trap \\
\texttt{101\_\allowbreak{}obby\_\allowbreak{}flatten\_\allowbreak{}segments} & \texttt{classic\_\allowbreak{}obby} & $\bullet$ & \,--\, & Flatten all obby segments within a consistent y axis \\
\texttt{102\_\allowbreak{}city\_\allowbreak{}spawn\_\allowbreak{}on\_\allowbreak{}tallest\_\allowbreak{}building} & \texttt{modern\_\allowbreak{}city} & $\bullet$ & \,--\, & Set the spawn for the players on top of the tallest building \\
\texttt{103\_\allowbreak{}city\_\allowbreak{}lights\_\allowbreak{}on\_\allowbreak{}off} & \texttt{modern\_\allowbreak{}city} & \,--\, & $\bullet$ & Turn the lights on/off based on time of day \\
\texttt{104\_\allowbreak{}lasertag\_\allowbreak{}mobile\_\allowbreak{}camera\_\allowbreak{}recoil} & \texttt{laser\_\allowbreak{}tag} & $\bullet$ & $\bullet$ & My camera recoil is too high when the user is playing from a mobile device. Change the recoil amount based on the user's platform. \\
\texttt{105\_\allowbreak{}lasertag\_\allowbreak{}friendly\_\allowbreak{}fire} & \texttt{laser\_\allowbreak{}tag} & $\bullet$ & $\bullet$ & Allow players to hit members of their own team / Allow friendly fire in this game. \\
\texttt{106\_\allowbreak{}lasertag\_\allowbreak{}weapon\_\allowbreak{}balance} & \texttt{laser\_\allowbreak{}tag} & $\bullet$ & $\bullet$ & Can you change the behavior of the two weapon types in my game? I want one of them to feel more like a shotgun, easier to hit but with lower damage, and the other one to feel more like a sniper rifle, harder to hit but with higher damage. \\
\texttt{107\_\allowbreak{}lasertag\_\allowbreak{}grenade\_\allowbreak{}weapon} & \texttt{laser\_\allowbreak{}tag} & $\bullet$ & $\bullet$ & Add a realistic grenade weapon as a third weapon option to this game. \\
\texttt{108\_\allowbreak{}racing\_\allowbreak{}fix\_\allowbreak{}getdriversinpart} & \texttt{racing} & $\bullet$ & $\bullet$ & Fix the error: getDriversInPart is not a valid member of ModuleScript "ServerScriptService.Racing.RaceManager" \\
\texttt{110\_\allowbreak{}racing\_\allowbreak{}car\_\allowbreak{}offtrack\_\allowbreak{}reset} & \texttt{racing} & \,--\, & $\bullet$ & Add a 'reset' functionality to the game that detects whether the car is upside down or is off-track for 5s. The reset functionality should teleport the car to the starting position on the track with the player. \\
\texttt{113\_\allowbreak{}racing\_\allowbreak{}extend\_\allowbreak{}laps} & \texttt{racing} & $\bullet$ & $\bullet$ & Make the races last longer. I want them to be 20 laps. \\
\texttt{115\_\allowbreak{}racing\_\allowbreak{}wrong\_\allowbreak{}direction\_\allowbreak{}warning} & \texttt{racing} & \,--\, & $\bullet$ & If the player is going in the wrong direction in the race, show a warning on their screen that says 'WRONG DIRECTION'. \\
\texttt{117\_\allowbreak{}fnf\_\allowbreak{}enter\_\allowbreak{}game\_\allowbreak{}ui\_\allowbreak{}menu} & \texttt{baseplate} & $\bullet$ & $\bullet$ & 1. Before entering the game, display a menu with: - "Play" button that leads to the scene. - "Freeplay" button to choose songs (only placeholder). - "Exit" button that closes the menu. 2. Add a logo that says "Friday Night Funkin' Roblox" \\
\texttt{118\_\allowbreak{}weapon\_\allowbreak{}spawn\_\allowbreak{}and\_\allowbreak{}pickup} & \texttt{baseplate} & \,--\, & $\bullet$ & Make the items pistol, axe, and knife spawn in various locations on the map, and ensure that the player can only pick up one of each. \\
\texttt{119\_\allowbreak{}lasertag\_\allowbreak{}add\_\allowbreak{}megablaster} & \texttt{laser\_\allowbreak{}tag} & $\bullet$ & $\bullet$ & Add a MegaBlaster that works in StarterPack. \\
\texttt{121\_\allowbreak{}monster\_\allowbreak{}chase\_\allowbreak{}at\_\allowbreak{}night} & \texttt{baseplate} & \,--\, & $\bullet$ & Make 'scary monsters' spawn at night, they will chase the player \\
\texttt{122\_\allowbreak{}animal\_\allowbreak{}item\_\allowbreak{}with\_\allowbreak{}rarity} & \texttt{baseplate} & \,--\, & $\bullet$ & I want 6 items of different rarity to spawn in the middle, each item must have a unique animal shape.

Rarity with random income between two values for each rarity: Common 60\% \$1-5, Rare 20\% \$6-50, Very Rare 11\% \$100-1,000, Mythic 7\% \$1,000-10,000, Legendary 1\% \$10,000-50,000, Ultra 0.9\% \$50,000-100,000, God 0.09\% \$100,000-300,000, Limited 0.01\% Random value between \$300,000-500,000.

6 items [...] \\
\end{longtable}
\end{footnotesize}
\end{center}

\begin{center}
\begin{footnotesize}
\begin{longtable}{@{}%
>{\raggedright\arraybackslash}p{0.190\linewidth}%
>{\raggedright\arraybackslash}p{0.120\linewidth}%
@{\hspace{5pt}}c@{\hspace{5pt}}c@{\hspace{7pt}}%
>{\raggedright\arraybackslash}p{0.440\linewidth}@{}}
\caption{The 29 debugging tasks. Each is a core task with a defect introduced and a prompt written from the point of view of a developer who has hit it, so the place file is shared with the core task it derives from.}\label{tab:catalog-debug}\\
\toprule
Task & Place & Scene & Play & Prompt \\
\midrule
\endfirsthead
\multicolumn{5}{@{}l}{\emph{Table~\ref{tab:catalog-debug}, continued.}}\\
\toprule
Task & Place & Scene & Play & Prompt \\
\midrule
\endhead
\bottomrule
\endfoot
\texttt{001\_\allowbreak{}make\_\allowbreak{}cars\_\allowbreak{}faster\_\allowbreak{}bug\_\allowbreak{}2} & \texttt{racing} & $\bullet$ & \,--\, & I was trying to make the cars faster, but now they don't spawn in the game at all and I'm getting errors. I think I might have moved the car prefab somewhere by accident. \\
\texttt{002\_\allowbreak{}emit\_\allowbreak{}white\_\allowbreak{}smoke\_\allowbreak{}bug\_\allowbreak{}1} & \texttt{surburban} & $\bullet$ & \,--\, & The chimneys are emitting black smoke instead of white smoke. It looks really weird and pollutes the neighborhood vibe. \\
\texttt{003\_\allowbreak{}make\_\allowbreak{}leaves\_\allowbreak{}fall\_\allowbreak{}colored\_\allowbreak{}bug\_\allowbreak{}1} & \texttt{village} & $\bullet$ & \,--\, & I tried to make the leaves on the trees change to autumn colors, but they all turned bright blue and green instead. That doesn't look like fall at all. \\
\texttt{004\_\allowbreak{}reduce\_\allowbreak{}car\_\allowbreak{}friction\_\allowbreak{}enable\_\allowbreak{}sliding\_\allowbreak{}bug\_\allowbreak{}1} & \texttt{racing} & $\bullet$ & \,--\, & I'm trying to run a script to make the cars drift more, but it's not working. The friction feels exactly the same. I think the script might not be finding the car parts correctly. \\
\texttt{004\_\allowbreak{}reduce\_\allowbreak{}car\_\allowbreak{}friction\_\allowbreak{}enable\_\allowbreak{}sliding\_\allowbreak{}bug\_\allowbreak{}2} & \texttt{racing} & $\bullet$ & \,--\, & I tried to make the cars more slippery so they could drift, but now they feel even more stuck to the road and difficult to turn. It feels like the friction went up instead of down. What's wrong? \\
\texttt{007\_\allowbreak{}change\_\allowbreak{}logo\_\allowbreak{}cube\_\allowbreak{}green\_\allowbreak{}bug\_\allowbreak{}1} & \texttt{platformer} & $\bullet$ & \,--\, & I tried to make the floating Roblox logo cube green, but it turned red instead. Can you fix it? \\
\texttt{007\_\allowbreak{}change\_\allowbreak{}logo\_\allowbreak{}cube\_\allowbreak{}green\_\allowbreak{}bug\_\allowbreak{}2} & \texttt{platformer} & $\bullet$ & \,--\, & I'm trying to change the color of the big floating Roblox logo cube to green, but it keeps turning red right after I change it. It seems like some other script is interfering and overriding the color. \\
\texttt{008\_\allowbreak{}spawn\_\allowbreak{}as\_\allowbreak{}r6\_\allowbreak{}bug\_\allowbreak{}1} & \texttt{baseplate} & $\bullet$ & $\bullet$ & Players are spawning as the newer R15 models, but we need them to be the classic R6 blocky characters for our game. \\
\texttt{010\_\allowbreak{}left\_\allowbreak{}shift\_\allowbreak{}sprint\_\allowbreak{}5s\_\allowbreak{}bug\_\allowbreak{}1} & \texttt{baseplate} & \,--\, & $\bullet$ & I tried setting up a sprint script, but whenever I hold down the Left Shift key, my character slows down to a crawl instead of speeding up. \\
\texttt{010\_\allowbreak{}left\_\allowbreak{}shift\_\allowbreak{}sprint\_\allowbreak{}5s\_\allowbreak{}bug\_\allowbreak{}2} & \texttt{baseplate} & \,--\, & $\bullet$ & When I hold Left Shift to sprint, it works, but if I let go of the key early, I keep sprinting for the full 5 seconds. I should stop sprinting as soon as I release the key. \\
\texttt{010\_\allowbreak{}left\_\allowbreak{}shift\_\allowbreak{}sprint\_\allowbreak{}5s\_\allowbreak{}bug\_\allowbreak{}3} & \texttt{baseplate} & \,--\, & $\bullet$ & When I hold the sprint key, it works, but it feels like the sprint lasts way too long. It's supposed to time out after 5 seconds, but it seems to go on for much longer than that before my character gets tired and slows down. \\
\texttt{011\_\allowbreak{}change\_\allowbreak{}height\_\allowbreak{}to\_\allowbreak{}1\_\allowbreak{}bug\_\allowbreak{}1} & \texttt{baseplate} & \,--\, & $\bullet$ & I wrote a script to make players 1 stud tall, but when I test it in Studio, my character is still the normal height. The script doesn't seem to be doing anything. \\
\texttt{011\_\allowbreak{}change\_\allowbreak{}height\_\allowbreak{}to\_\allowbreak{}1\_\allowbreak{}bug\_\allowbreak{}2} & \texttt{baseplate} & \,--\, & $\bullet$ & I wrote a script to make player characters 1 stud tall when they spawn, but it's not working. They're still spawning at their default height. \\
\texttt{012\_\allowbreak{}remove\_\allowbreak{}legs\_\allowbreak{}bug\_\allowbreak{}1} & \texttt{baseplate} & \,--\, & $\bullet$ & When my character spawns, they still have legs. The script is supposed to remove them but it doesn't seem to be working. \\
\texttt{012\_\allowbreak{}remove\_\allowbreak{}legs\_\allowbreak{}bug\_\allowbreak{}2} & \texttt{baseplate} & \,--\, & $\bullet$ & When I join the game, my character's legs are supposed to be removed, but they're still there. The script doesn't seem to be working. \\
\texttt{012\_\allowbreak{}remove\_\allowbreak{}legs\_\allowbreak{}bug\_\allowbreak{}3} & \texttt{baseplate} & \,--\, & $\bullet$ & I wrote a script to remove player legs when they spawn, but it doesn't work for my character when I first load into the game. It only seems to work for players who join the server after I'm already there. \\
\texttt{013\_\allowbreak{}inf\_\allowbreak{}cube\_\allowbreak{}fall\_\allowbreak{}bug\_\allowbreak{}1} & \texttt{baseplate} & \,--\, & $\bullet$ & I wrote a script that's supposed to make a cube fall at the center of the map every second, but nothing is happening. No parts are appearing at all. \\
\texttt{013\_\allowbreak{}inf\_\allowbreak{}cube\_\allowbreak{}fall\_\allowbreak{}bug\_\allowbreak{}2} & \texttt{baseplate} & \,--\, & $\bullet$ & I wrote a script to make a cube fall from the sky every second, but it seems to be happening much slower than that. I'm only seeing a new cube every few seconds. \\
\texttt{017\_\allowbreak{}make\_\allowbreak{}traffic\_\allowbreak{}light\_\allowbreak{}bug\_\allowbreak{}1} & \texttt{baseplate} & \,--\, & $\bullet$ & I'm trying to make a traffic light for my game, but it's not working. The lights are just stuck on and never change color or cycle. \\
\texttt{017\_\allowbreak{}make\_\allowbreak{}traffic\_\allowbreak{}light\_\allowbreak{}bug\_\allowbreak{}2} & \texttt{baseplate} & \,--\, & $\bullet$ & I added the traffic light model to my game, but it's just stuck. The lights never turn on or change color. \\
\texttt{017\_\allowbreak{}make\_\allowbreak{}traffic\_\allowbreak{}light\_\allowbreak{}bug\_\allowbreak{}3} & \texttt{baseplate} & \,--\, & $\bullet$ & I added a script to control the traffic light, but it's not working. The lights are stuck and don't change color. \\
\texttt{020\_\allowbreak{}gravity\_\allowbreak{}well\_\allowbreak{}bug\_\allowbreak{}1} & \texttt{baseplate} & \,--\, & $\bullet$ & I set up the gravity well part, but when players touch it, they get slammed into the ground instead of floating up. \\
\texttt{021\_\allowbreak{}gem\_\allowbreak{}orbiting\_\allowbreak{}part\_\allowbreak{}bug\_\allowbreak{}1} & \texttt{baseplate} & $\bullet$ & $\bullet$ & I tried to make a gem that orbits a part, but the gem isn't showing up in the game at all. It's completely invisible and doesn't seem to exist in the Workspace. \\
\texttt{022\_\allowbreak{}add\_\allowbreak{}world\_\allowbreak{}time\_\allowbreak{}bug\_\allowbreak{}1} & \texttt{baseplate} & \,--\, & $\bullet$ & The time of day is stuck and the sun isn't moving. I thought I added the script for the day-night cycle, but it doesn't seem to be working. \\
\texttt{022\_\allowbreak{}add\_\allowbreak{}world\_\allowbreak{}time\_\allowbreak{}bug\_\allowbreak{}2} & \texttt{baseplate} & \,--\, & $\bullet$ & The day/night cycle seems to be running way too fast. It feels like time is speeding up more than it should. \\
\texttt{022\_\allowbreak{}add\_\allowbreak{}world\_\allowbreak{}time\_\allowbreak{}bug\_\allowbreak{}3} & \texttt{baseplate} & \,--\, & $\bullet$ & I wrote a script to make the day/night cycle faster, but the time isn't changing at all. The sun is just stuck in the same spot. \\
\texttt{023\_\allowbreak{}music\_\allowbreak{}playing\_\allowbreak{}part\_\allowbreak{}bug\_\allowbreak{}1} & \texttt{baseplate} & \,--\, & $\bullet$ & I set up a trigger part that's supposed to play music when a player walks into it, but the music never starts. I've checked that the sound and the part exist, but nothing happens when my character touches the part. \\
\texttt{023\_\allowbreak{}music\_\allowbreak{}playing\_\allowbreak{}part\_\allowbreak{}bug\_\allowbreak{}2} & \texttt{baseplate} & \,--\, & $\bullet$ & I created a trigger part that's supposed to play music when a player walks into it, but the music never starts. The part and the sound are both in the Workspace, but nothing happens. \\
\texttt{023\_\allowbreak{}music\_\allowbreak{}playing\_\allowbreak{}part\_\allowbreak{}bug\_\allowbreak{}3} & \texttt{baseplate} & \,--\, & $\bullet$ & I set up a trigger part to play music when a player enters it, but it seems to be working backwards. The music stops when I enter the zone and only starts playing again when I leave. What's wrong with my script? \\
\end{longtable}
\end{footnotesize}
\end{center}

\end{document}